\documentclass[conference]{IEEEtran}
\IEEEoverridecommandlockouts
\usepackage{adjustbox}
\usepackage{lineno}
\usepackage{algorithm}
\usepackage{algorithmic}
\usepackage{multirow}
\usepackage[tight,footnotesize]{subfigure}
\usepackage{epsfig}

\usepackage{amssymb}
\usepackage{footmisc}
\usepackage{footnpag}
\usepackage{amsmath}
\usepackage{paralist}
\usepackage{color}
\usepackage{perpage}
\usepackage{verbatim}
\usepackage{enumitem}
\usepackage{tikz}
\usepackage{listings}
\usepackage{graphicx}
\usepackage{soul}
\usepackage{fancyhdr}
\usepackage[inkscapeformat=png]{svg}
\usepackage{stfloats}
\usepackage{balance}

\def\BibTeX{{\rm B\kern-.05em{\sc i\kern-.025em b}\kern-.08em
    T\kern-.1667em\lower.7ex\hbox{E}\kern-.125emX}}
    
\setlist[itemize]{topsep=\parskip}

\begin{document}

\newcommand\xin[1]{{\color{cyan}{Xin: #1}}}
\newcommand\sheng[2]{{\color{red}{Sheng: #1}}}

\pagestyle{plain}

\title{SRN-SZ: Deep Leaning-Based Scientific Error-bounded Lossy Compression with Super-resolution Neural Networks}

\author{\IEEEauthorblockN{Jinyang Liu,\IEEEauthorrefmark{1}
Sheng Di,\IEEEauthorrefmark{2}
Sian Jin, \IEEEauthorrefmark{6}
Kai Zhao,\IEEEauthorrefmark{5}
Xin Liang,\IEEEauthorrefmark{3}
Zizhong Chen,\IEEEauthorrefmark{1}
Franck Cappello\IEEEauthorrefmark{2}\IEEEauthorrefmark{4}}
\IEEEauthorblockA{\IEEEauthorrefmark{1}University of California, Riverside, CA, USA}
\IEEEauthorblockA{\IEEEauthorrefmark{2}Argonne National Laboratory, Lemont, IL, USA}

\IEEEauthorblockA{\IEEEauthorrefmark{3}
University of Kentucky, Lexington, KY, USA}
\IEEEauthorblockA{\IEEEauthorrefmark{4}
University of Illinois at Urbana-Champaign, Urbana, IL, USA}
\IEEEauthorblockA{\IEEEauthorrefmark{5}
Florida State University, Tallahassee, FL, USA}
\IEEEauthorblockA{\IEEEauthorrefmark{6}
Indiana University Bloomington, Bloomington, IN, USA}
jliu447@ucr.edu, sdi1@anl.gov, sianjin@iu.edu, kzhao@cs.fsu.edu, xliang@cs.uky.edu, chen@cs.ucr.edu, cappello@mcs.anl.gov
\thanks{Corresponding author: Sheng Di, Mathematics and Computer Science Division, Argonne National Laboratory, 9700 Cass Avenue, Lemont, IL 60439, USA}
}

\maketitle

\thispagestyle{fancy}
\lhead{}
\rhead{}
\chead{}
\rfoot{}
\cfoot{}
\renewcommand{\headrulewidth}{0pt} \renewcommand{\footrulewidth}{0pt}

\begin{abstract}
The fast growth of computational power and scales of modern super-computing systems have raised great challenges for the management of exascale scientific data. To maintain the usability of scientific data, error-bound lossy compression is proposed and developed as an essential technique for the size reduction of scientific data with constrained data distortion. Among the diverse datasets generated by various scientific simulations, certain datasets cannot be effectively compressed by existing error-bounded lossy compressors with traditional techniques. The recent success of Artificial Intelligence has inspired several researchers to integrate neural networks into error-bounded lossy compressors. However, those works still suffer from limited compression ratios and/or extremely low efficiencies. To address those issues and improve the compression on the hard-to-compress datasets, in this paper, we propose SRN-SZ, which is a deep learning-based scientific error-bounded lossy compressor leveraging the hierarchical data grid expansion paradigm implemented by super-resolution neural networks. SRN-SZ applies the most advanced super-resolution network HAT for its compression, which is free of time-costing per-data training. In experiments compared with various state-of-the-art compressors, SRN-SZ achieves up to 75\% compression ratio improvements under the same error bound and up to 80\% compression ratio improvements under the same PSNR than the second-best compressor.

\end{abstract}


\section{Introduction}
\label{sec:introduction}

The rapid growth of computing power of worldwide exascale supercomputers has enabled the scientific applications running on them to intensely enlarge their scales and outputs. Nevertheless, the data storage capacity and memory bandwidth of those machines have not developed fast enough to catch up with the increasingly huge amount of data generated by those applications, bringing rising requirements for advanced data reduction techniques to efficiently store, transfer, and analyze those data. To this end, error-bounded lossy compression has been recognized as the most proper strategy to manage extremely large amounts of scientific data. Compared to the lossless compression techniques which can only provide around halved compressed size, it can reduce the data size to 10\%, 1\%, or even 0.1\% of the original size. Unlike many existing lossy compressors (such as the JPEG compressor for image data) that do not constrain the point-wise data error, error-bounded lossy compression can control the point-wise data distortion upon the user's requirements. Therefore, error-bounded lossy compression is of great significance for boosting the utility of scientific data.

Existing state-of-the-art scientific error-bounded lossy compressors with diverse compression ratios and speeds, such as SZ3 \cite{szinterp,sz3}, ZFP \cite{zfp}, and SPERR \cite{SPERR}, have shown advantages in variant practical use cases. However, despite the success existing error-bounded lossy compressors have achieved, their limitations persist. Among the diverse archetypes of existing compressors, their compressions on certain datasets are still apparently under-optimized, suffering from low compression ratios, which is still an ongoing challenge for error-bounded lossy compression research.

Inspired by the great breakthroughs in the Artificial Intelligence field, several attempts have been made to leverage neural networks in error-bounded lossy compression. Autoencoder-based AE-SZ \cite{ae-sz} and Coordinate network-based CoordNet \cite{han2022coordnet} are two typical examples. Those deep learning-based compressors may provide well-optimized compression ratios in certain cases, but their limitations are still obvious. The Coordinate network-based compressors \cite{han2022coordnet,huang2022compressing,lu2021compressive} suffer from extremely low compression efficiencies as they need to train a new network separately for each input. Although autoencoder-based compressors such as \cite{ae-sz,hayne2021using} can leverage pre-trained networks to avoid per-input training, their compression ratios cannot overperform SZ3 in most cases \cite{ae-sz}.

In order to address the issues of optimizing the hard-to-compress data compression and overcoming the limitations of deep learning-based error-bounded lossy compression, in this paper, we proposed SRN-SZ, which is a grand new deep learning-based error-bounded lossy compression framework. The core innovation of SRN-SZ is that it abstracts the compression and decompression processes of scientific data grids into a hierarchical paradigm of data grid super-resolution, which is the first work of integrating the super-resolution neural network into the error-bounded lossy compressor to the best of our knowledge. Compared with the autoencoders and coordinate networks, the super-resolution networks have two-fold advantages: Unlike coordinate networks, they can be pre-trained before the practical compression tasks. At the same time, they do not generate any latent information that is required to be stored for compression as the autoencoders. Benefiting from those advantages, SRN-SZ achieves acceptable efficiencies and further improved compression ratios over the state-of-the-art error-bounded lossy compressors on multiple hard-to-compress datasets. 

The contributions of our paper are detailed as follows:
\begin{itemize}
    \item We propose a new scientific error-bounded lossy compressor SRN-SZ, in which the compression is performed by hierarchical data grid expansion implemented with a hybrid of super-resolution networks and interpolations.
    \item Leveraging the Hybrid Attention Transformer (HAT) network, we designed a specialized training pipeline with several adaptive techniques to optimize the super-resolution quality of scientific data.
    \item We carry out systematical evaluations with SRN-SZ and 5 other state-of-the-art scientific error-bounded lossy compressors on various scientific datasets from different domains. According to the experimental results, SRN-SZ has achieved up to 75\% compression ratio improvements under the same error bound and up to 80\% compression ratio improvements under the same PSNR.
\end{itemize}

The rest of this paper is organized as follows: In, Section \ref{sec:related}, we discuss related works. Section \ref{sec:problemform} presents the research problem formulation and backgrounds. The overall framework of SRN-SZ is demonstrated in Section \ref{sec:design}. The compression pipeline and network training pipeline of SRN-SZ are separately proposed in Section \ref{sec:pipeline} and Section \ref{sec:training}. In section \ref{sec:evaluation}, the evaluation results are provided and analyzed. Section \ref{sec:conclusion} concludes this work and discusses future work. 
\section{Related Work}
\label{sec:related}

In this section, we discuss the related works in 3 categories: Traditional scientific error-bounded lossy compression, deep learning-based scientific lossy compression, and super-resolution neural networks.
\subsection{Traditional Scientific Error-bounded Lossy Compression}
Traditional scientific error-bounded lossy compressors can be classified into prediction-based, transform-based, and dimension-reduction-based. The prediction-based compressors utilize different data prediction techniques for the compression, such as linear regression (SZ2 \cite{Xin-bigdata18}) and interpolations (SZ3 \cite{szinterp} and QoZ \cite{qoz}). Transform-based compressors decorrelate the input data by data transformation techniques so that the transformed data (a.k.a., coefficients) turn out to be much easier to compress than the original dataset; then it compresses the efficient domain to get a high compression ratio. Typical examples include ZFP \cite{zfp} leveraging orthogonal discrete transform and SPERR \cite{SPERR} integrating CDF 9/7 wavelet transform. With dimension reduction techniques such as (high-order) singular vector decomposition (SVD), dimension-reduction-based compressors such as TTHRESH \cite{ballester2019tthresh} can perform the data compression very effectively. Besides the CPU-based compressors, several GPU-specialized error-bounded lossy compressors have also been developed and proposed for better parallelization and throughput. Typical examples are CuSZ \cite{cusz,cusz+} and FZ-GPU \cite{FZGPU}.

\subsection{Deep Learning-based Scientific Lossy Compression}
The great success of the recent research of Artificial Intelligence techniques started boosting the development of several other relevant research fields, including the scientific error-bounded lossy compression. Several research works that leverage deep neural networks in error-bounded lossy compression have been proposed \cite{ae-sz,han2022coordnet,huang2022compressing,lu2021compressive,hayne2021using}. There are mainly 2 archetypes: autoencoder-based compressors which store the autoencoder-encoded latent vectors for compression, and coordinate network-based compressors which train networks online for each input to map the data coordinates to data values. For autoencoder-based compressors, AE-SZ is an example of integrating Slice-Wasserstein autoencoders, and Hayne et al. \cite{hayne2021using} leverages a double-level autoencoder for compressing 2D data. Examples of coordinate network-based compressors include NeurComp \cite{lu2021compressive}, CoordNet \cite{han2022coordnet} and \cite{huang2022compressing}.
\subsection{Super-resolution Neural Networks}

Following the SRCNN \cite{dong2014learning} which introduced a Convolutional neural network model to the image super-resolution tasks, a large number of convolutional neural network models \cite{ahn2018fast,lim2017enhanced,zhang2018image} 
have been proposed for the super-resolutions. Because of the development of Transformer \cite{transformer} and its adaption to Computer Vision tasks \cite{vit,liu2021swin,wu2021cvt}, vision-transformer-based neural networks like \cite{chen2021pre,liang2021swinir,hat} have achieved state-of-the-art performance on the image super-resolution task. Among those works, HAT \cite{hat} is the most impressive one as it has the widest scope of feature extraction for reconstructing each data point with a carefully designed hybrid attention model and achieves state-of-the-art performance.

\section{Problem Formulation and Backgrounds}
\label{sec:problemform}
\subsection{Mathematical Formulations for Error-bounded Lossy Data Compression}
In this subsection, we propose several key mathematical definitions and the mathematical formulation of our research target for this paper.  

\subsubsection{Compression ratio and bit rate} Compression ratio is defined by the input data size divided by the compressed data size. Specifically, for input data $X$ and compressed data $Z$, compression ratio $\rho$ is:
 \vspace{-2mm}
\begin{equation}
\label{eq:cr}
  \rho=\frac{|X|}{|Z|}
  \vspace{-1mm}
\end{equation}
According to Eq. \ref{eq:cr}, a higher compression ratio means better (smaller) compressed size, and vice versa. In the visualization of experimental results, researchers often plot curves with another metric closely related to the compression ratio, namely the bit rate. Bit rate is defined by the average number of bytes used in the compressed data to store each data element for the input data, which can be expressed as (denote bit rate by $b$):

\begin{equation}
\label{eq:br}
  b=\frac{sizeof(x)}{|Z|}
  \vspace{-1mm}
\end{equation}
in which $x$ is an element of the input $X$, and sizeof() returns the byte size. Since the bit rate is reciprocal to the compression ratio, a lower bit rate is better.

\subsubsection{PSNR} PSNR (Peak Signal-to-Noise Ratio) is one of the most important data distortion metrics for evaluating the quality of the decompressed data from the lossy compression. it is defined as follows: 
\def\formulaPSNR{
P\hspace{-0.3mm}S\hspace{-0.3mm}N\hspace{-0.3mm}R = 20\log_{10}{vrange(X)} \hspace{-0.3mm}-\hspace{-0.3mm} 10\hspace{-0.3mm}\log_{10}{mse(X,\hspace{-0.3mm}X')} \hspace{-0.1mm}
}
\vspace{-2mm}
\begin{equation}
\label{eq:psnr}
  \formulaPSNR
  \vspace{-1mm},
\end{equation}
\noindent
where $X$ is the input data and $X'$ is the decompressed data. vrange() calculates the value range of one data array, and \emph{mse} refers to the mean-squared error. Fixing the input data (and also the data range), a smaller mean-squared error will lead to higher PSNR, therefore higher PSNR means higher precision of the decompressed data. 
\subsubsection{Research target} The objective of SRN-SZ is to optimize the compression process with regard to a certain optimization target: maximizing compression PSNR under each certain compression ratio. Mathematically speaking, given the input data $X$, compressed data $Z$, decompression output $X'$, error bound $e$, and the target compression ratio $T$, we will optimize the compressor $C$ and decompressor $D$ of SRN-SZ via the following optimization problem ($Z = C(X)$ and $ X' = D(Z)$):
\def\formulaEBFORM{
\begin{array}{l}
 maximize\hspace{2mm}PSNR(X,X')  \\ 
 s.t. \hspace{4mm} \frac{|X|}{|Z|}=T \\
 \hspace{8mm}\hspace{1mm}|x_i  - x_i'| \leq e, \ \forall x_i \in X.
 \end{array}
}
\vspace{-2mm}
\begin{equation}
\label{eq:opt}
  \formulaEBFORM  ,
\end{equation}

In this paper, we propose a deep learning-based compressor, leveraging the super-resolution neural network for the optimization of  Eq. \ref{eq:opt}. 

\subsection{Challenge for Error-bounded Lossy Compression: Low-compression-ratio Datasets}
Recently proposed scientific error-bound lossy compressors have succeeded in outperforming the old state-of-the-art compressors dramatically. Compared with the historical SZ 2.1 \cite{Xin-bigdata18}, SZ3 \cite{sz3} has improved the compressor ratio by up to 460\% \cite{szinterp} under the same data distortion. With higher computational costs, wavelet-based compressors such as SPERR \cite{SPERR} may have doubled or even tripled compression ratios compared with SZ3. 

\begin{figure}[ht] 
\vspace{1mm}
\centering
\hspace{-10mm}
\subfigure[{Miranda}]
{
\raisebox{-1cm}{\includegraphics[scale=0.25]{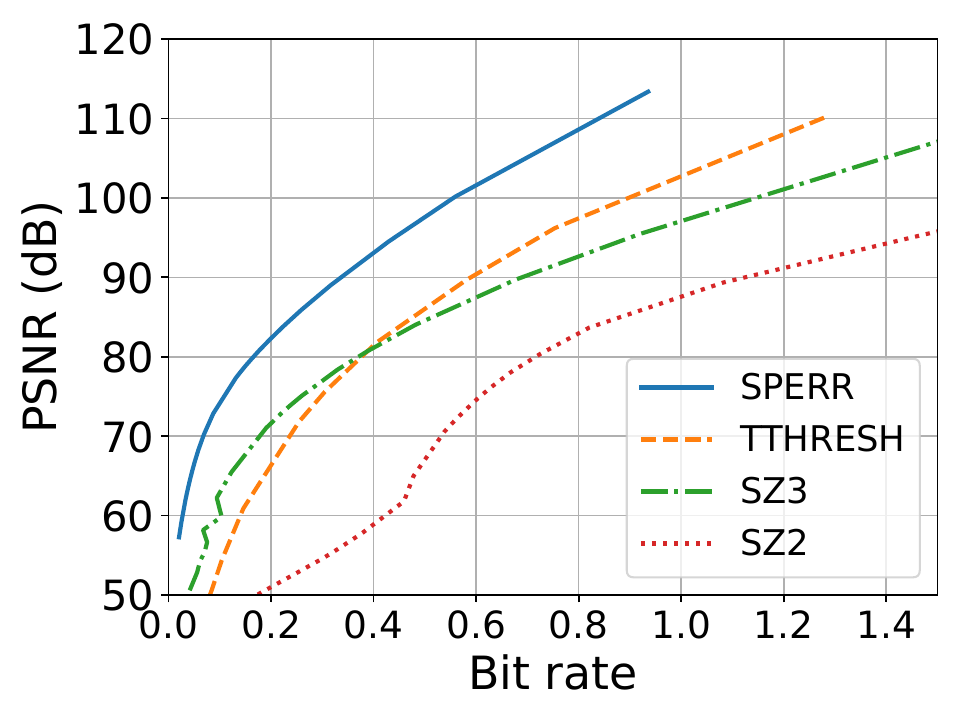}}
}
\hspace{-4mm}
\subfigure[{CESM-CLDHGH}]
{
\raisebox{-1cm}{\includegraphics[scale=0.25]{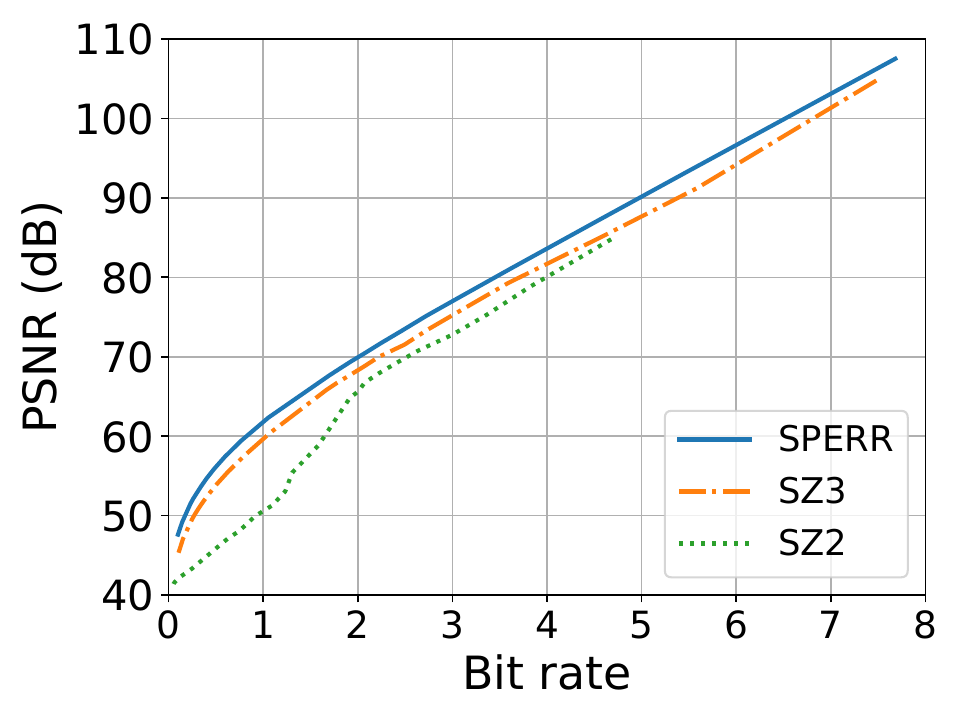}}
}
\hspace{-10mm}

\vspace{0mm}
\hspace{-10mm}
\subfigure[{NYX-Dark Matter Density}]
{
\raisebox{-1cm}{\includegraphics[scale=0.25]{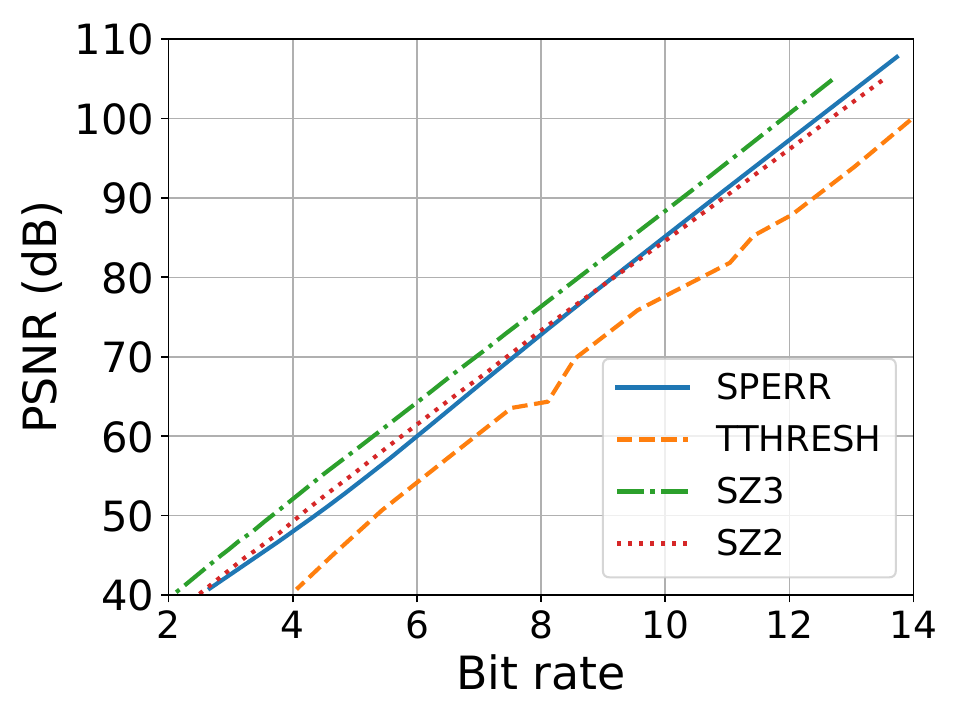}}
}
\hspace{-4mm}
\subfigure[{Hurricane-QRAIN}]
{
\raisebox{-1cm}{\includegraphics[scale=0.25]{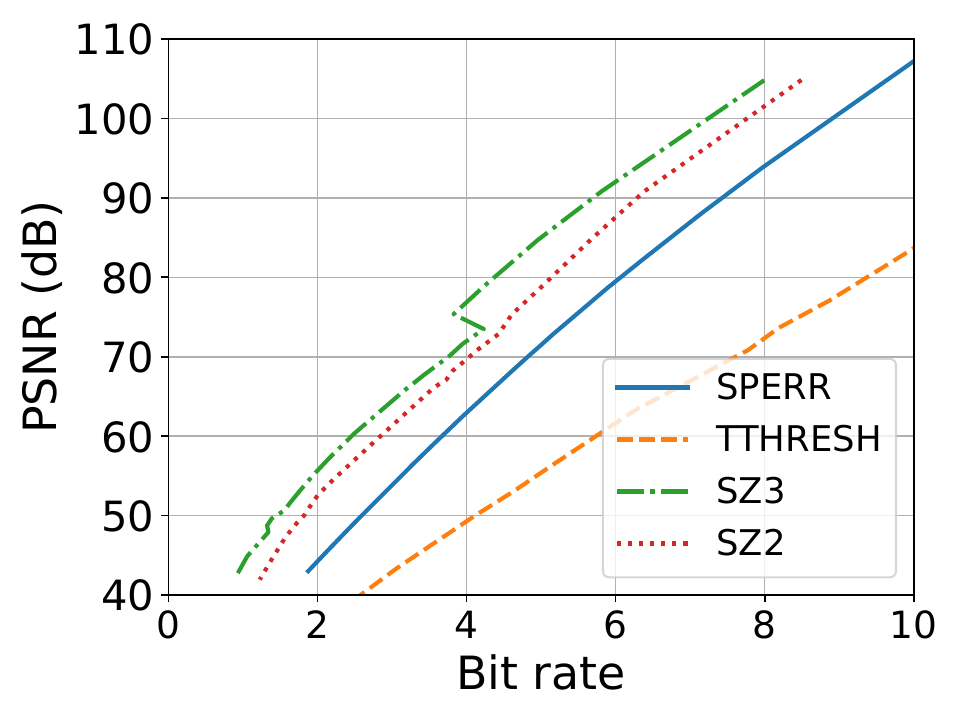}}%
}
\hspace{-10mm}
\vspace{-1.5mm}

\caption{Rate-distortion (PSNR) of several existing error-bounded compressors. }
\label{fig:analysis-rate-psnr}
\vspace{-1mm}
\end{figure}

However, those exciting improvements in compression ratios are just concentrated on datasets that generally project relatively high compression ratios (e.g. over 100). In other words, the recent proposed works with advanced data compression techniques fail to improve the compression for datasets with relatively low compression ratios to similar extents as they have done in high-ratio cases. In Figure \ref{fig:analysis-rate-psnr}, we present the bit rate-PSNR curves from the compression of 4 scientific datasets with the representative existing error-bounded lossy compressors: prediction-based SZ2 \cite{Xin-bigdata18} and SZ3 \cite{szinterp,sz3}, SVD-based TTHRESH \cite{ballester2019tthresh}, and wavelet transform-based SPERR \cite{SPERR} (the compression result of TTHRESH is not shown in Figure \ref{fig:analysis-rate-psnr} (b) as TTHRESH does not support 2D data input). For datasets like the Miranda \cite{miranda} (Figure \ref{fig:analysis-rate-psnr} (a)). SZ3 has boosted the compression ratio of SZ2 by over 100\%, and SPERR further achieves 2x-3x of the compression ratio over SZ3. However, on other datasets, those 4 compressors have relatively low compression ratios. On certain datasets such as NYX-Dark Matter Density and Hurricane-QRain (Figure \ref{fig:analysis-rate-psnr} (c) and (d)), the SPERR and TTHRESH have lower compression ratios than SZ3 does, though they are designed with more complicated data processing techniques and much higher computational costs.

It is worth noting that the low-compression-ratio data snapshots are actually the bottleneck of compression effectiveness because their compressed data size will obviously occupy a very large portion of all data fields (having diverse characteristics) in a single dataset. For example, compressing 100TB of data with a compression ratio of 100 will generate 1TB of compressed data, which means that we can at most save the space of 1TB when optimizing the compression. Nevertheless, if the original data has the same size of 100TB but only has a potential compression ratio of 5 (20TB compressed data), merely improving the compression ratio by 10\% will lead to around 1.8TB storage cost reduction. Therefore, overcoming the limitation of existing compressors on low-compression-ratio data will be significant for optimizing the overall compression process for a large variety of scientific simulation datasets.

\section{SRN-SZ Design Overview}
\label{sec:design}
\begin{figure}[ht]
  \centering
  \vspace{1mm}
  \raisebox{-1mm}{\includegraphics[scale=0.48]{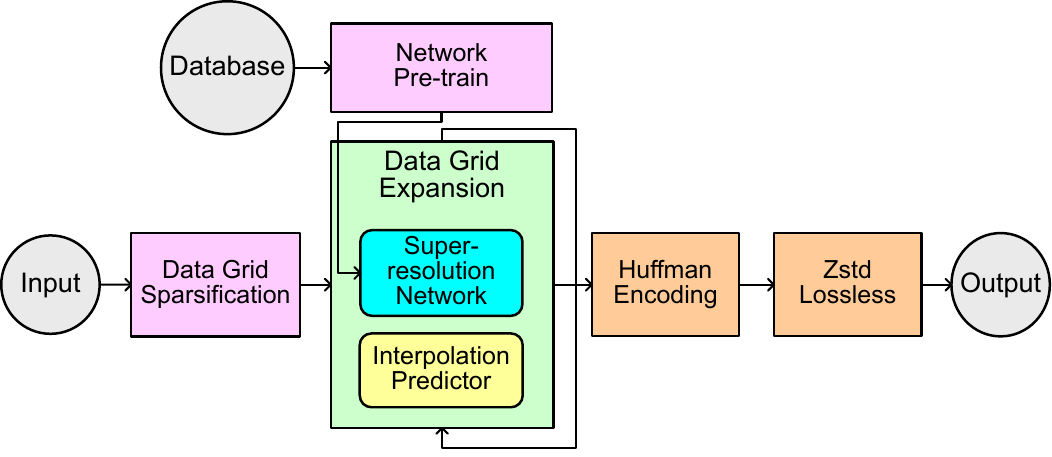}}
  \vspace{-1mm}
  \caption{SRN-SZ compression framework}
  \label{fig:framework}
\end{figure}

We would like to propose our SRN-SZ, which is a deep learning-based error-bounded lossy compressor, and is based on a modular compression framework that integrates a hybrid data reconstruction model with both interpolators and super-resolution neural networks. As shown in Figure \ref{fig:framework}, the compression framework of SRN-SZ consists of 4 modules: Data grid sparsification, data grid expansion, Huffman encoding, and Zstd lossless compression. Moreover, the super-resolution neural networks are first pre-trained with a large-size dataset assorted from the scientific database and then fine-tuned with domain-specific datasets before being leveraged in the data grid expansion module of SRN-SZ. In the compression process of SRN-SZ, it first extracts a sparse data grid from the original data input, next, this sparse data grid is expanded step by step with super-resolution networks and interpolators, eventually to a lossy reconstruction of the full-size input grid. Compared to existing deep learning-based compressors which leverage autoencoder-like networks \cite{ae-sz,liu2021high} to generate compact representations or coordinate networks \cite{han2022coordnet,huang2022compressing,lu2021compressive} mapping data point indices to data values, SRN-SZ has the advantages of both free from the storage cost for the compact representations (required by autoencoders) and per-input network training (required by coordinate networks).

We demonstrate the detailed compression algorithm of SRN-SZ in Algorithm \ref{alg:srnsz}. Lines 1-2 correspond to data grid sparsification, Lines 3-10 correspond to data grid expansion, and Lines 11-12 correspond to Huffman encoding and Zstd lossless compression. To bound the point-wise compression error, the linear quantization is involved in the data grid expansion module, and for clearness of demonstration, it is not displayed in Figure \ref{fig:framework}. %

\begin{algorithm}
\footnotesize
\caption{SRN-SZ Compression Algorithm}
\label{alg:srnsz}
\renewcommand{\algorithmiccomment}[1]{/*#1*/}
\begin{flushleft}
\textbf{Input:} Input data $D$, error-bound $e$, grid sparsification rate $r$, minimum SRN size $s$\\ 
\textbf{Output:} Compressed data $Z$
\end{flushleft}

\begin{algorithmic}[1]
\STATE Sparsify $D$ into $D_0$ with rate $r$. Save $D_0$ losslessly \COMMENT{Data grid sparsification.}
\STATE Set current reconstructed data grid $D^{'}\leftarrow D_0$, Quantized errors $Q\leftarrow \{\}$
\WHILE{$size(D^{'})!=size(D)$}

\IF {$size(D^{'}) \le s$}
\STATE{$D^{'},q=Interp\_and\_Quantize(D,D^{'},e)$}\COMMENT{Expand $D^{'}$ with interpolation.}
\ELSE 
\STATE {$D^{'},q=HAT\_and\_Quantize(D,D^{'},e)$} \COMMENT{Expand $D^{'}$ with HAT network.} 
\ENDIF
\STATE {$Q \leftarrow Q \bigcup q$}. \COMMENT{Merge newly acquired quantized errors $q$.}
\ENDWHILE
\STATE $H$ $\leftarrow$ Huffman\_Encode($Q$). \COMMENT{Huffman encoding}
\STATE $Z$ $\leftarrow$ \emph{Zstd}($H,D_0$). \COMMENT{Zstd compression}
\end{algorithmic}
\end{algorithm}

\section{SRN-SZ Compression Pipeline}
\label{sec:pipeline}
In this section, we describe the steps in the SRN-SZ Compression pipeline in detail. Since the encoding and lossless modules of SRN-SZ are the same as the ones in SZ3 and QoZ \cite{szinterp,sz3,qoz}, in the following subsections, we will mainly discuss the data grid sparsification and data grid expansion. 
\subsection{Data Grid Sparsification}
\begin{figure}[ht]
  \centering
  \vspace{1mm}
  \raisebox{-1mm}{\includegraphics[scale=0.75]{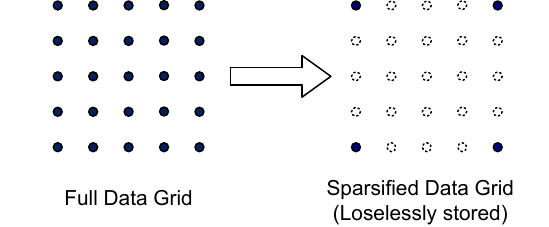}}
  \vspace{-1mm}
  \caption{Data grid sparsification}
  \label{fig:sparsification}
  \vspace{-2mm}
\end{figure}

Having shown advantages in MGARD \cite{MGARD,liang2021mgard+}, SZ3 \cite{szinterp,sz3}, and QoZ \cite{qoz}, SRN-SZ adopts a level-wise hierarchical data grid reconstruction paradigm for its compression process. It starts from a sparse data grid sampled from the original input dataset. An example of 2D input data is shown in Figure \ref{fig:sparsification}: certain data points are uniformly sampled from the full data grid with a fixed stride. Those sampled data points in a sparsified data grid will be losslessly saved and the rest data points will be reconstructed in the data grid expansion process. The reason SRN-SZ losslessly saves the sparsified grid instead of directly reconstructing a lossy version of it from scratch as SZ3 does is analyzed below. According to the comparison between evaluations of SZ3 and QoZ \cite{qoz}, for the hierarchical level-wise data reconstruction, an accurate base is essential for preserving the high reconstruction quality of the data points, meanwhile only introducing negligible overhead storage space overhead. To balance the compression ratio loss and data reconstruction accuracy, we conducted some tests and then specified the dimension-wise rate of data grid sparsification as $\frac{1}{32}$, i.e., reduce the data grid to $\frac{1}{32}$ along each dimension and then save the sparsified grid for the data grid expansion.       
\subsection{Data Grid Expansion}
\label{sec:expansion}
\begin{figure}[ht]
  \centering
  \vspace{1mm}
  \raisebox{-1mm}{\includegraphics[scale=0.75]{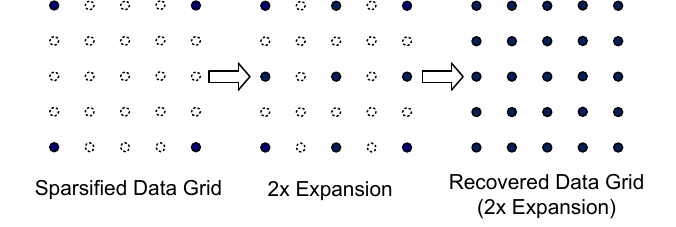}}
  \vspace{-1mm}
  \caption{Data grid expansion}
  \label{fig:expansion}
\end{figure}

Based on the sparsified data grid, the data grid expansion (i.e. reconstruction) process is involved in both the compression and decompression of SRN-SZ. In the compression, the data grid expansion is executed for acquiring the reconstruction errors of data points, and then those errors are quantized and encoded serving as the correction offsets in the decompression. Moreover, During both the compression and decompression process of SRN-SZ, the super-resolution and error-quantization in compression (or error correction in decompression) are executed alternately, which can maximally preserve the accuracy of data grid expansion. As presented in Figure \ref{fig:expansion}, the data grid expansion is performed iteratively step by step, until the whole data grid has been reconstructed. In each step, the reconstructed data grid is expanded by 2x along each dimension, therefore its implementation is compatible with both the deep learning-based super-resolution neural networks and the traditional interpolation methods.
\subsubsection{HAT super-resolution network}
\begin{figure}[ht]
  \centering
  \vspace{1mm}
  \raisebox{-1mm}{\includegraphics[scale=0.5]{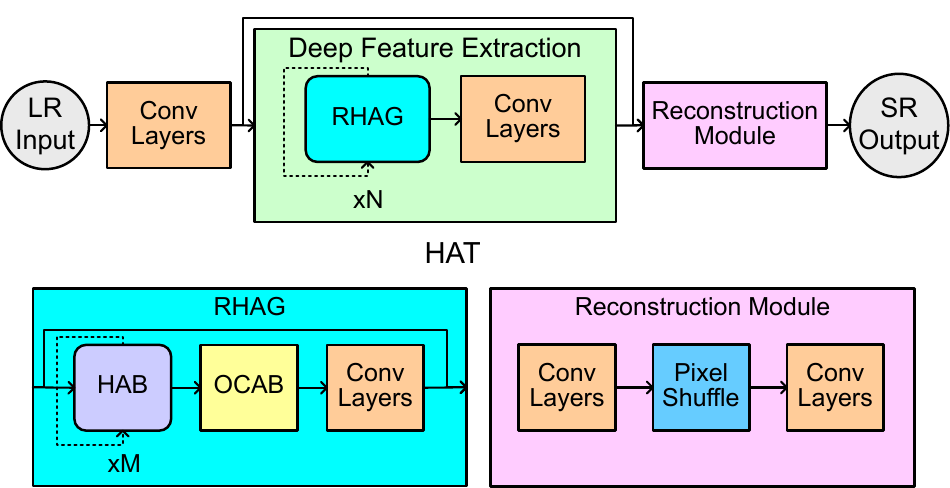}}
  \vspace{-1mm}
      \caption{HAT network}
  \label{fig:hat}
\end{figure}

 Super-resolution network is the most important data grid expansion technique in SRN-SZ as it is always applied on the last iteration step of data grid expansion, which contains the reconstruction for most of the data points in the input data (about 75\% for 2D case and about 87.5\% for 3D case). The network SRN-SZ leveraged is the HAT (Hybrid Attention Transformer) network \cite{hat}, which is a very recent proposed work for image super-resolution and has been proven to be state-of-the-art. The network architecture of HAT is illustrated in Figure \ref{fig:hat}. Developed from \cite{liang2021swinir,zhang2018image}, HAT is a very-deep residual \cite{he2016deep} neural network with transformers \cite{transformer} as its basic components. HAT has 3 main modules: the initial convolutional layers for shallow feature extraction, the deep feature extraction module integrated with residual hybrid attention groups (RHAG), and a reconstruction module leveraging the Pixel Shuffle technique \cite{shi2016real}. The RHAG blocks in the HAT network can be broken down into HAB (hybrid attention block), OCAB (overlapping cross-attention block), and convolutional layers. For more details of the HAT network, we refer readers to read \cite{hat}. The main advantage of HAT is that according to the analysis presented in \cite{hat}, the design of HAT empowers it to make use of a large region of data points for computing each value in its super-resolution output. Therefore, both local and global data patterns can be well utilized in the super-resolution process.
 
\begin{figure}[ht]
  \centering
  \vspace{1mm}
  \raisebox{-1mm}{\includegraphics[scale=0.25]{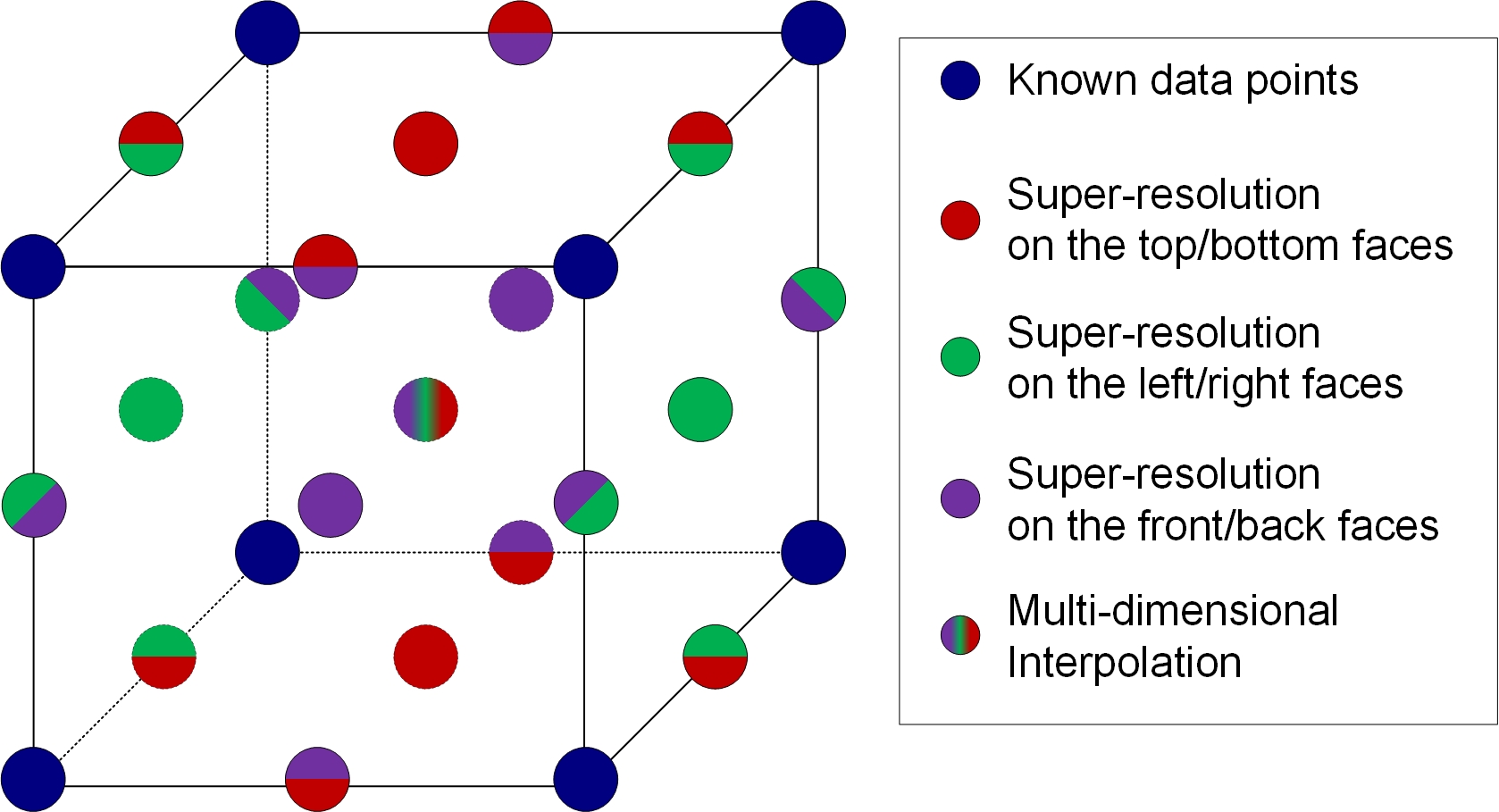}}
  \vspace{-1mm}
      \caption{3D super-resolution with 2D slices}
  \label{fig:2dslices}
\end{figure}

\begin{figure}[ht]
  \centering
  \vspace{1mm}
  \raisebox{-1mm}{\includegraphics[scale=0.6]{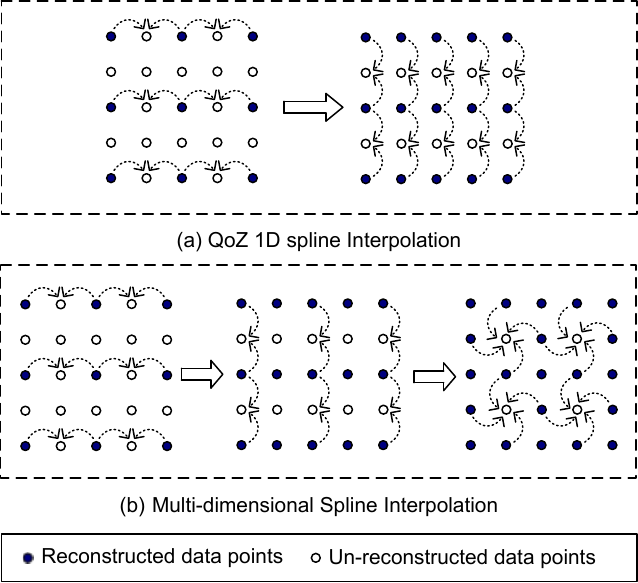}}
  \vspace{-1mm}
      \caption{Interpolations in SRN-SZ}
  \label{fig:interp}
\end{figure}

 Although HAT was originally designed for the super-resolution of natural images, we managed to adapt it to the scientific data grid expansion process in SRN-SZ. Feeding an intermediate data grid with size X x Y (or X x Y x Z) into HAT, SRN-SZ uses the super-resolution output of size 2X x 2Y (or 2X x 2Y x 2Z) from HAT as the data grid expansion result in one step. Some key points in bridging the scientific data and the HAT network are: First, the input and output channels in HAT have been modified from 3 to 1. Second, the input data grid is normalized to 0-1 before being fed into the network. Last, for 3D data inputs, 2D HAT models can still be used, but the inputs are preprocessed into 2D slices (along all the 3 dimensions) instead of 3D blocks. The reason SRN-SZ applies 2D networks for 3D data is that 3D HAT models suffer from extremely high computational time costs for training and inference, presenting unacceptable flexibility and scalability. Figure \ref{fig:2dslices} presents the details of performing 3D super-resolution with those 2D slices. Specifically, with a partially reconstructed 3D data grid (blue points), SRN-SZ performs super-resolution on it with the HAT network in 3 different directions: on top/bottom faces (red points), on left/right faces (green points), and on front/back faces (purple points). The super-resolution results on the edges are the average of 2 directions, and the point on the cubic center is reconstructed by a multi-dimensional spline interpolation, which is introduced in \cite{HPEZ} and will be detailed in the next subsection and Figure \ref{fig:interp} (b).

\subsubsection{interpolation-based data predictor}
We have observed that, when the reconstructing data grid has a small size, the super-resolution network can not work well. Therefore, on some initial steps of data grid expansion in which the current data grid is smaller than a threshold (with a dimension shorter than 64), the traditional QoZ-based interpolation \cite{qoz} is leveraged for the grid expansion which can auto-tune the best-fit interpolation configurations and error bounds. In addition to the QoZ interpolation, following the design proposed by \cite{HPEZ}, SRN-SZ also leverages several advanced interpolation designs such as multi-dimensional spline interpolation. Figure \ref{fig:interp} presents and compares these two interpolation methods, and SRN-SZ will dynamically select the interpolation method for each interpolation level. This adaptive selection design improves both the efficiency of SRN-SZ and the reconstruction quality in the early steps of data grid expansion. 

\section{SRN-SZ Network Training}
\label{sec:training}
\begin{figure*}[ht]
  \centering
  \vspace{1mm}
  \raisebox{-1mm}{\includegraphics[scale=0.55]{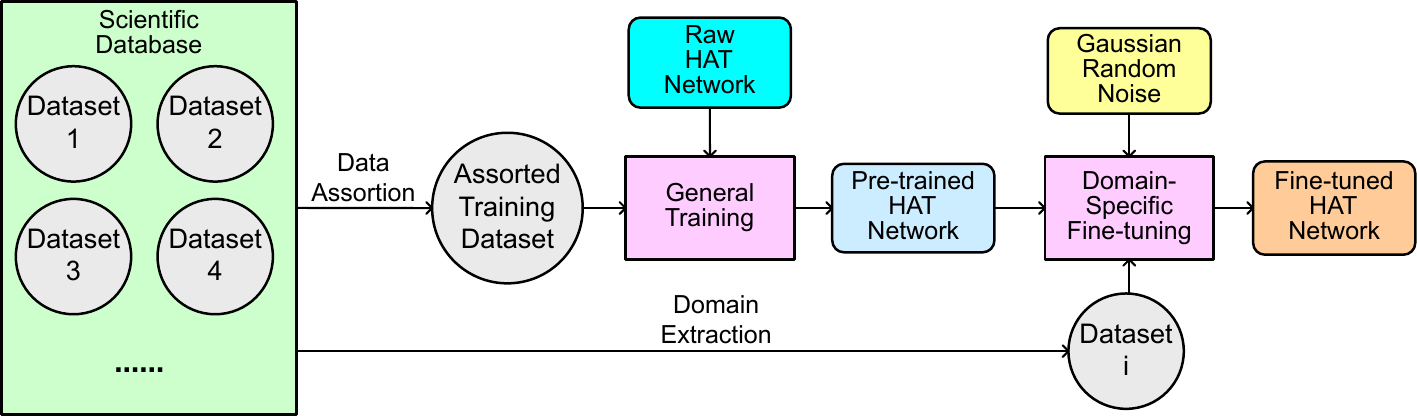}}
  \vspace{-1mm}
  \caption{SRN-SZ network training pipeline}
  \label{fig:train}
\end{figure*}

The super-resolution quality of the HAT network plays the most important role in optimizing the compression ratio with controlled data distortion in SRN-SZ, and the core of optimizing the super-resolution quality of the HAT is its training process. The HAT networks in SRN-SZ are pre-trained offline both with an assorted dataset and domain-specific datasets. This design contributes to the flexibility and adaptability of SRN-SZ. Several strategies have been proposed for optimizing the training of the HAT networks in SRN-SZ. Figure \ref{fig:train} proposes our designed HAT network training pipeline for SRN-SZ. In the pipeline, each network is trained for two rounds: the general training from scratch and the domain-specific training for fine-tuning. The following subsections describe the key design of this pipeline.
\subsection{Training data collection and preprocessing}
We have collected training data snapshots from a variety of well-known scientific simulations, including CESM-ATM \cite{cesm}, RTM \cite{geodriveFirstBreak2020}, OCEAN, Miranda \cite{miranda}, JHTDB \cite{jhtdb}, Hurricane-ISABEL \cite{hurricane}, SCALE-LetKF \cite{scale-letkf}, NYX \cite{nyx}, and so on. The full list of the scientific simulations used by SRN-SZ for the HAT network training is shown in Table \ref{tab:traindata}. With those assorted data snapshots, we first decompose 3D data arrays into 2D data slices, next normalize them to [0,1] range, then split all over-sized (over 480x480) slices into smaller slices (480x480) according to the setting in \cite{hat}. When yielding the training data batches, the low-resolution, and high-resolution image pairs are randomly cropped from those slices. The widely-used image data augmentation methods like random flip and rotation are excluded from SRN-SZ network training as we observe that those data augmentation strategies will harm the quality of super-resolution with test data. This assorted and pre-processed dataset will be used for general pre-training of the HAT network from scratch.
\begin{table}[ht]
    \vspace{-1mm}
    \centering
    \caption{Information of the scientific simulations for training data of SRN-SZ}  \vspace{-2mm}  
\resizebox{0.99\columnwidth}{!}{  
    \begin{tabular}{|c|c|c|c|}
    \hline
    App. & Dimensions &Per snapshot size& Domain\\
    \hline
    CESM-ATM &2D& 1800$\times$3600&Climate \\
    \hline
    Hurricane & 3D&100$\times$500$\times$500 & Weather\\
    \hline
    JHTDB &3D&  512$\times$512$\times$512& Turbulence \\
    \hline
    Miranda &3D&  256$\times$384$\times$384& Turbulence \\
    \hline
     NYX & 3D&512$\times$512$\times$512& Cosmology\\
    \hline
    OCEAN & 2D&2400$\times$3600 & Oceanology\\
    \hline
    RTM &3D& 449$\times$449$\times$235&Seismic Wave\\
    \hline
    Scale-LETKF &3D& 98$\times$1200$\times$1200&Climate\\
    \hline
    \end{tabular}}
    \label{tab:traindata}
\end{table}

\subsection{Domain-specific fine-tuning}
\label{sec:ft}
Datasets from different scientific domains and simulations would present diverse patterns and characteristics. To make the trained network better adapt to more varied inputs, we fine-tune the super-resolution for certain scientific simulations that are being intensively and consistently used for research and analysis. To address this issue, we develop a domain-specific fine-tuning in SRN-SZ. After an initial training phase with the assorted database, SRN-SZ picks up several additional data snapshots generated by those simulations and then fine-tunes the network separately with each simulation data. In this way, SRN-SZ can achieve improved compression ratios on multiple widely used scientific data simulation datasets. We will compare the rate-distortion of SRN-SZ between applying the domain-specific fine-tuning or not in Section \ref{sec:abla}.

\subsection{Denoise training with Gaussian random noise}
\label{sec:denoise}
As discussed in Section \ref{sec:expansion}, the data grid to be expanded in SRN-SZ is a lossy sample from the original data input. At the same time, we will need the super-resolution of it to fit the original input as much as possible. To simulate this process in the training of the HAT networks in SRN-SZ for better super-resolution results, we propose denoise training in SRN-SZ. Specifically, instead of simply using full data grids and the corresponding down-sampled data grids as the training data pairs, SRN-SZ adds Gaussian noise to the down-sampled data grids before feeding them into the network in the training phase. In this way, the trained network will be capable of de-noising the input for more accurate super-resolution outputs. Moreover, we observe that training networks with intense noises will damage their effectiveness on low error-bound cases, so we separately train 3 base networks with different intensities of noises: strong noise (with stand derivation of 1\% of data range), weak noise (with stand derivation of 0.1\% of data range), and no noise. Those networks will correspondingly serve for different compression cases: high error bounds (larger than 1e-2), medium error bounds (1e-4 to 1e-2), and low error bounds (smaller than 1e-4).

\section{Performance Evaluation}
\label{sec:evaluation}

In this section, we describe the setup of our experiments and then present the experimental results together with our analysis. We evaluate the newly proposed SRN-SZ and compare it with five other state-of-the-art error-bounded lossy compressors \cite{sz3,SPERR,qoz,ballester2019tthresh,liu2023faz}.

\subsection{Experimental Setup}

\subsubsection{Experimental environment and datasets}
Our experiments are conducted on the Argonne Bebop supercomputer (for CPU-based tests) and the ALCF Theta supercomputer (for GPU-based tests). On the Bebop machine, we used its nodes of the bdwall partition, having an Intel Xeon E5-2695v4 CPU with 64 CPU cores and a total of 128GB of DRAM on each. On the Theta machine, each GPU node of it has 8 NVIDIA TESLA A100 GPUs.

We select 6 data fields from 4 real-world scientific applications in diverse scientific domains. Those datasets are frequently used for evaluating scientific error-bounded lossy compression \cite{sdrb}.
We detail the information about the datasets and the fields in Table~\ref{tab:dataset information}. As suggested by domain scientists, some fields of the datasets listed above are transformed to their logarithmic domain for better visualization. For fairness of evaluation, the data snapshots used for the evaluations are never contained in the assorted training dataset and their corresponding fine-tuning datasets. However, for optimizing the compression, some data snapshots in the same data field (but from different runs of the application or from different time steps) are used for training (especially for fine-tuning).
\begin{table}[ht]
    \vspace{-1mm}
    \centering
    \caption{Information of the datasets in experiments}  \vspace{-2mm}  
\resizebox{0.99\columnwidth}{!}{  
    \begin{tabular}{|c|c|c|c|c|}
    \hline
    Name&\# fields& Dimensions &  Domain\\
    \hline
    CESM-ATM & CLDHGH, FREQSH & 1800$\times$3600&Climate\\
    \hline
    Ocean & TMXL & 2400$\times$3600 &Oceanology\\
    \hline
    NYX & Dark Matter Density & 512$\times$512$\times$512& Cosmology\\
    \hline
    Hurricane & QRain, QGraup & 100$\times$500$\times$500 & Weather\\
    \hline
    \end{tabular}}
    \label{tab:dataset information}
\end{table}

\subsubsection{Comparison of lossy compressors in evaluation}

In the experiments, SRN-SZ is evaluated together with five other state-of-the-art lossy compressors. Among those, 4 are the traditional error-bounded lossy compressors: SZ3 \cite{sz3}, QoZ \cite{qoz}, SPERR \cite{SPERR}, and FAZ \cite{liu2023faz}. Another one is the deep learning-based AE-SZ \cite{ae-sz}, which was verified in \cite{ae-sz} to be one of the most effective autoencoder-based error-bounded lossy compressors. We do not perform comparison experiments with coordinate-network-based compressors due to the reason that they suffer from very low compression speed (much slower than SRN-SZ) as they need to perform a network training process for each single compression task \cite{han2022coordnet,huang2022compressing,lu2021compressive}.

\subsubsection{Network training configurations}

For the training of HAT networks in SRN-SZ, we apply the network structure and training configurations described in \cite{hat}. In each training phase (including general training and domain-specific fine-tuning), we train the network on 8 GPUs in 200,000 iterations with a mini-batch size of 32. The initial learning rate is 2e-4 and will be halved on step [100K,160K,180K,190K]. For the network training and compression of AE-SZ, we follow the configurations described in \cite{ae-sz}.


\subsubsection{Evaluation Metrics}

In the compression experiments, we adopted the value-range-based error bound mode (denoted as $\epsilon$) being equivalent to the absolute error bound (denoted as $e$) with the relationship of $e$ = $\epsilon \cdot value\_range$. The evaluation results are based on the following key metrics: 
\begin{itemize}
    \item Decompression error verification: Verify that the decompression errors are strictly error-bounded.
    \item Compression ratio (CR) under the same error bound: Compression ratio is the metric mostly cared for by the users, for fair comparison, the compression ratios under fixed error bounds are presented. 
    \item \textit{Rate-PSNR plots}: Plot curves for compressors with the compression bit rate and decompression PSNR.
    \item {Visualization with the same CR: Comparing the visual qualities of the reconstructed data from different compressors based on the same CR.}
    \item {Ablation Study}: Verify the effectiveness of each SRN-SZ design component separately.
\end{itemize}

\subsection{Evaluation Results and Analysis}

\subsubsection{Verification of compression errors versus error bound}

\begin{figure}[ht] 
\centering
\hspace{-12mm}
\subfigure[{Hurricane-QRAIN}]
{
\raisebox{-1cm}{\includegraphics[scale=0.28]{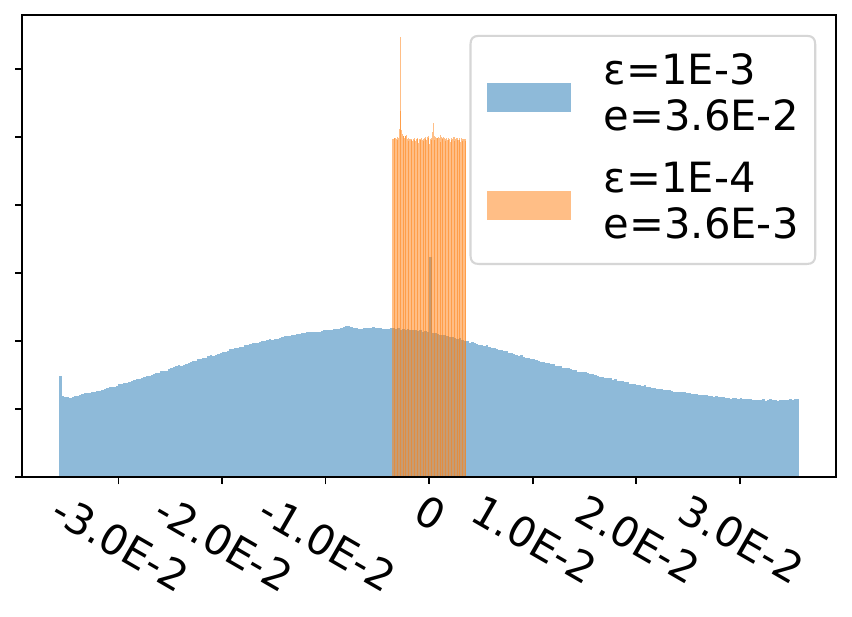}}
}%
\hspace{-3mm}
\subfigure[{Hurricane-QGRAUP}]
{
\raisebox{-1cm}{\includegraphics[scale=0.28]{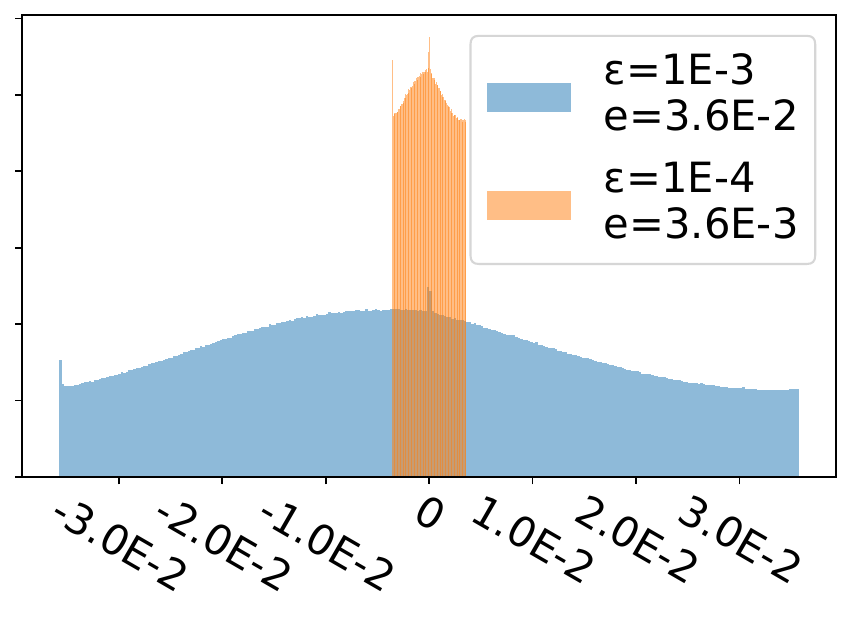}}%
}
\hspace{-12mm}
\vspace{-1mm}

\caption{Histograms of decompression errors from SRN-SZ}
\label{fig:err-verification}
\end{figure}

First of all, we verify that the decompression errors from SRN-SZ have strictly been constrained within the error bounds. To this end, we plot the histograms of decompression errors for each compression task, and two of them (on the QRAIN and QGRAUP fields of the Hurricane-ISABEL dataset) are presented in Figure \ref{fig:err-verification}. It can be clearly observed that the decompression errors of SRN-SZ always respect the error bound ($e$) in all cases with no out-of-bound abnormalities of point-wise decompression error. Having examined the error-bounded feature of SRN-SZ, in the following subsections, we will test, present, and analyze the compression ratios and qualities of SRN-SZ.

\subsubsection{Compression ratio under the same error bounds}

The compression ratios of all lossy compressors under the same certain error bounds (1e-3, 1e-4, and 1e-5) are presented in Table \ref{tab:compress ratio comparison}. An interesting fact is that, although proposed later than SZ3, some new compressors (QoZ, SPERR, and FAZ) have not raised the compression ratios well on the tested datasets. In contrast, SRN-SZ has quite improved the compression ratios of error-bounded lossy compressors on almost all of the tested compression cases, over a variety of datasets and error bounds. Particularly, under the error bound of 1e-4 SRN-SZ achieves a 75\% compression ratio improvement over the second-best QoZ on the CLDHGH field of the CESM-ATM dataset, and under the error bound of 1e-3 SRN-SZ achieves a 44\% compression ratio improvement on the FREQSH field of it. On other datasets, SRN-SZ can also get 3\% to 20\% compression ratio improvements. Last, compared with other deep learning-based compressors, SRN-SZ has outperformed AE-SZ in an overall assessment.

\begin{table}[ht]
\centering
\footnotesize
  \caption {Compression Ratio Comparison Based on the Same Error Bound} 
  \vspace{-2mm}
  \label{tab:compress ratio comparison} 
  \begin{adjustbox}{width=\columnwidth}
 \begin{tabular}{|c|c|c|c|c|c|c|c|c|}
\hline
\multirow{2}{*}{\textbf{Dataset}}                                                                      & \multirow{2}{*}{\textbf{$\epsilon$}} & \textbf{SZ}  & \multirow{2}{*}{\textbf{QoZ}} & \multirow{2}{*}{\textbf{SPERR}} & \textbf{AE-} & \multirow{2}{*}{\textbf{FAZ}} & \textbf{SRN-} & \textbf{Improve} \\
                                                                                                       &                                   & \textbf{3.1} &                               &                                 & \textbf{SZ}  &                               & \textbf{SZ}   & \textbf{(\%)}    \\ \hline
\multirow{3}{*}{\textbf{\begin{tabular}[c]{@{}c@{}}CESM\\       CLDHGH\end{tabular}}}                  & 1E-3                              & 19.2         & 18.8                          & 18.9                            & 16.8         & 16.0                          & \textbf{32.6} & 69.8             \\ \cline{2-9} 
                                                                                                       & 1E-4                              & 7.0          & 7.1                           & 7.0                             & 6.9          & 6.3                           & \textbf{12.4} & 74.6             \\ \cline{2-9} 
                                                                                                       & 1E-5                              & 4.3          & 4.1                           & 4.2                             & 4.1          & 3.8                           & \textbf{6.0}  & 39.5             \\ \hline
\multirow{3}{*}{\textbf{\begin{tabular}[c]{@{}c@{}}CESM\\      FREQSH\end{tabular}}}                   & 1E-3                              & 16.4         & 17.2                          & 16.3                            & 16.3         & 14.0                          & \textbf{24.7} & 43.6             \\ \cline{2-9} 
                                                                                                       & 1E-4                              & 6.4          & 6.6                           & 6.5                             & 6.5          & 5.9                           & \textbf{10.4} & 57.6             \\ \cline{2-9} 
                                                                                                       & 1E-5                              & 4.2          & 3.9                           & 4.0                             & 4.0          & 3.7                           & \textbf{5.1}  & 21.4             \\ \hline
\multirow{3}{*}{\textbf{\begin{tabular}[c]{@{}c@{}}Ocean\\      TMXL\end{tabular}}}                    & 1E-3                              & 25.3         & 24.9                          & 21.7                            & 23.4         & 15.5                          & \textbf{29.4} & 16.2             \\ \cline{2-9} 
                                                                                                       & 1E-4                              & 11.2         & 10.6                          & 9.7                             & 11.6         & 7.3                           & \textbf{12.1} & 4.3              \\ \cline{2-9} 
                                                                                                       & 1E-5                              & 6.5          & 6.6                           & 6.1                             & 7.0          & 4.7                           & \textbf{7.0}  & 0.7              \\ \hline
\multirow{3}{*}{\textbf{\begin{tabular}[c]{@{}c@{}}NYX\\      DarkMatter\\      Density\end{tabular}}} & 1E-3                              & 5.2          & 5.3                           & 4.5                             & 5.1          & 4.3                           & \textbf{5.9}  & 11.3             \\ \cline{2-9} 
                                                                                                       & 1E-4                              & 3.4          & 3.4                           & 3.1                             & 3.3          & 3.0                           & \textbf{3.7}  & 8.8              \\ \cline{2-9} 
                                                                                                       & 1E-5                              & 2.5          & 2.5                           & 2.3                             & 2.4          & 2.2                           & \textbf{2.6}  & 4.0              \\ \hline
\multirow{3}{*}{\textbf{\begin{tabular}[c]{@{}c@{}}Hurricane\\      QRAIN\end{tabular}}}               & 1E-3                              & 10.0         & 10.3                          & 6.9                             & 10.3         & 10.1                          & \textbf{11.2} & 8.7              \\ \cline{2-9} 
                                                                                                       & 1E-4                              & \textbf{6.5} & 5.3                           & 4.5                             & 5.8          & 4.2                           & 6.4           & -1.5             \\ \cline{2-9} 
                                                                                                       & 1E-5                              & 4.0          & 3.4                           & 3.2                             & 3.8          & 3.0                           & \textbf{4.1}  & 2.5              \\ \hline
\multirow{3}{*}{\textbf{\begin{tabular}[c]{@{}c@{}}Hurricane\\      QGRAUP\end{tabular}}}              & 1E-3                              & 11.2         & 11.2                          & 7.7                             & 11.0         & 11.2                          & \textbf{12.4} & 10.7             \\ \cline{2-9} 
                                                                                                       & 1E-4                              & \textbf{6.6} & 5.5                           & 4.8                             & 6.2          & 4.7                           & 6.0           & -9.1             \\ \cline{2-9} 
                                                                                                       & 1E-5                              & 4.0          & 3.5                           & 3.3                             & 3.9          & 3.3                           & \textbf{4.3}  & 7.5              \\ \hline
\end{tabular}
\end{adjustbox}
\end{table}

\subsubsection{Rate distortion evaluation}

Next, we present and analyze the rate-distortion evaluation of SRN-SZ and other state-of-the-art error-bounded lossy compressors.

\begin{figure}[ht] 
\centering
\hspace{-10mm}
\subfigure[{CESM-CLDHGH}]
{
\raisebox{-1cm}{\includegraphics[scale=0.26]{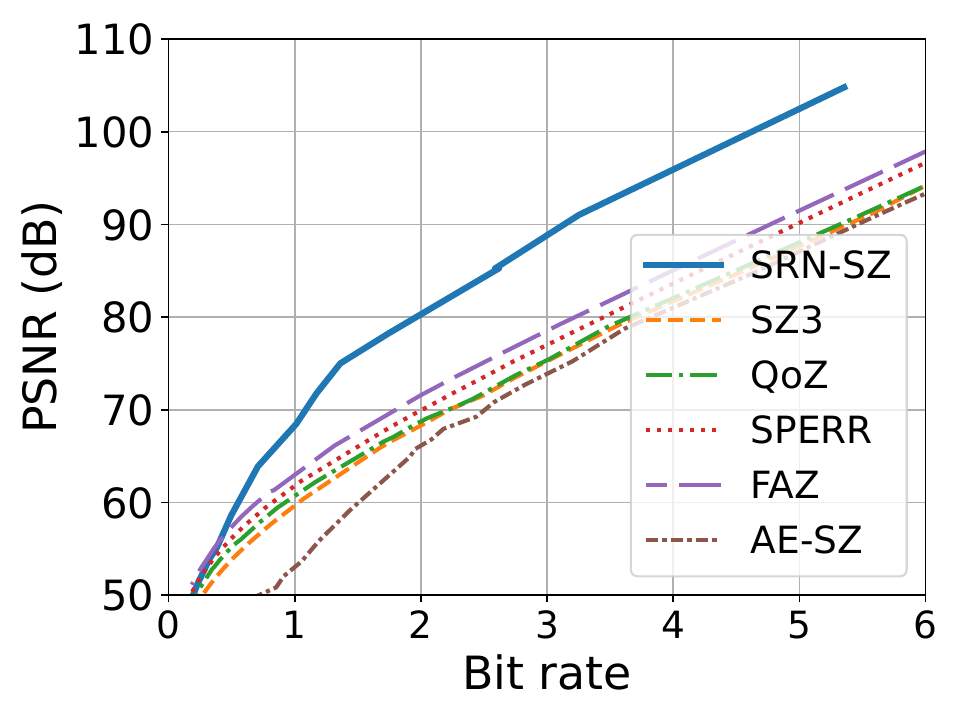}}
}
\hspace{-6mm}
\subfigure[{CESM-FREQSH}]
{
\raisebox{-1cm}{\includegraphics[scale=0.26]{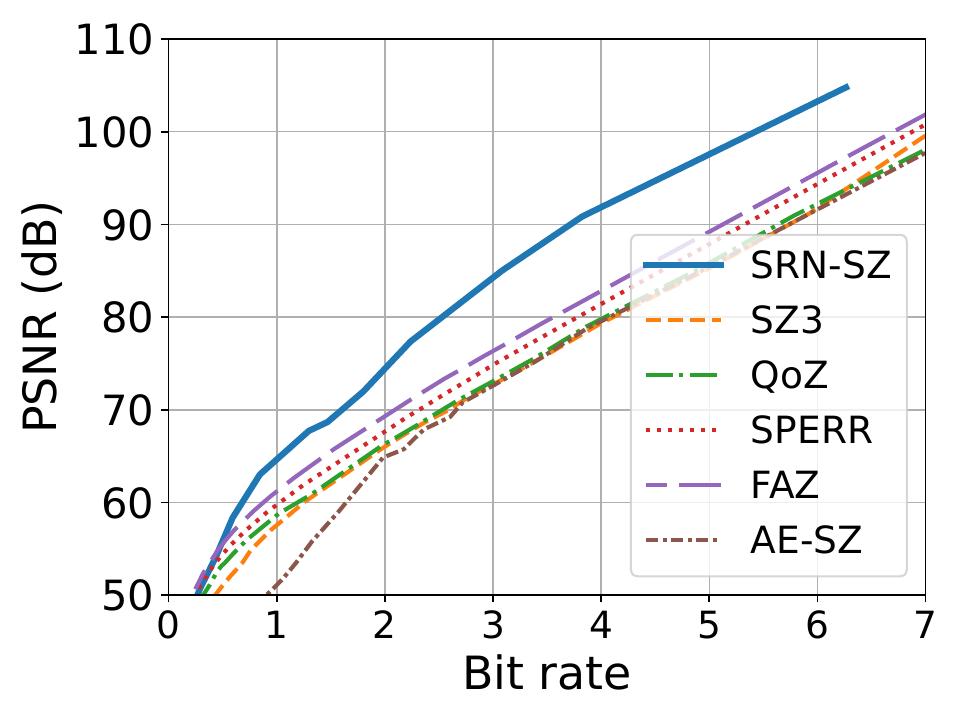}}
}
\hspace{-10mm}
\vspace{-1mm}

\hspace{-10mm}
\subfigure[{Ocean-TMXL}]
{
\raisebox{-1cm}{\includegraphics[scale=0.26]{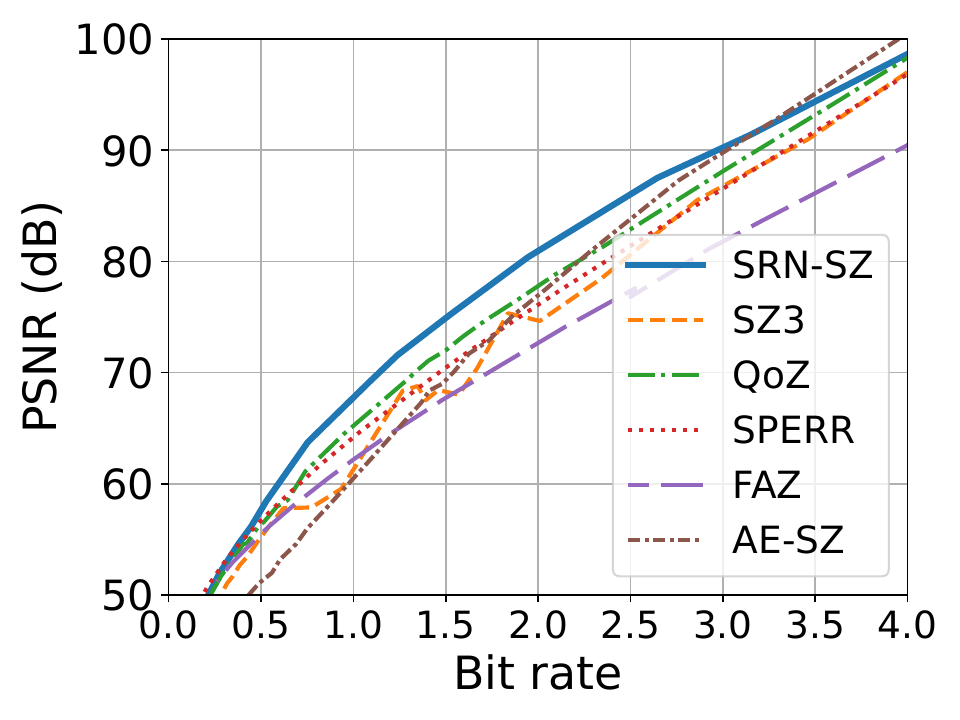}}
}
\hspace{-6mm}
\subfigure[{NYX-Dark Matter Density}]
{
\raisebox{-1cm}{\includegraphics[scale=0.26]{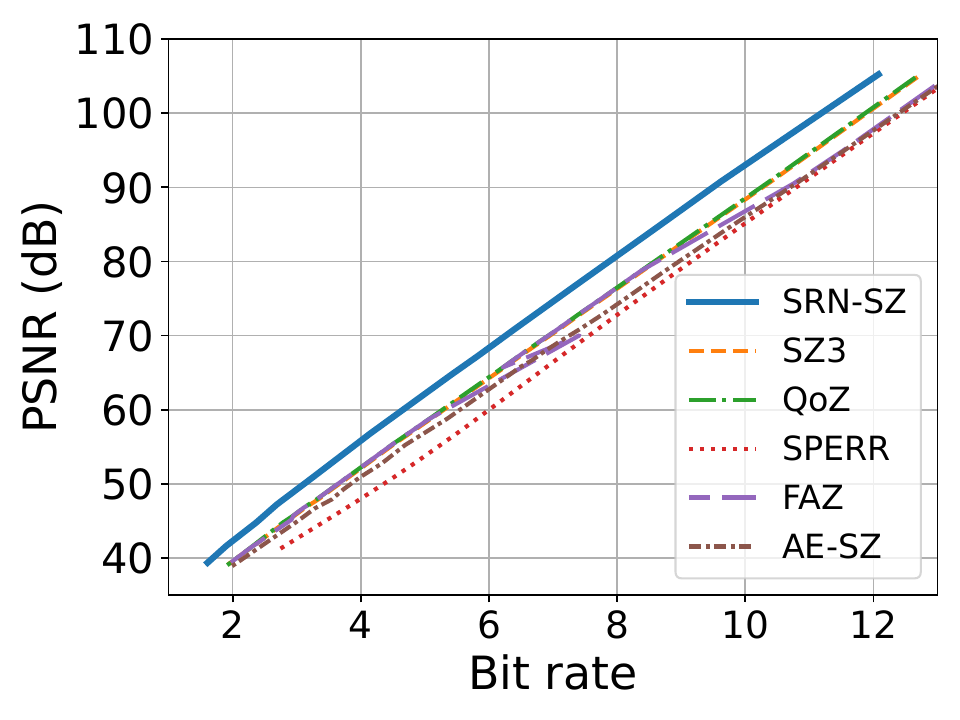}}%
}
\hspace{-10mm}
\vspace{-1mm}

\hspace{-10mm}
\subfigure[{Hurricane-QRAIN}]
{
\raisebox{-1cm}{\includegraphics[scale=0.26]{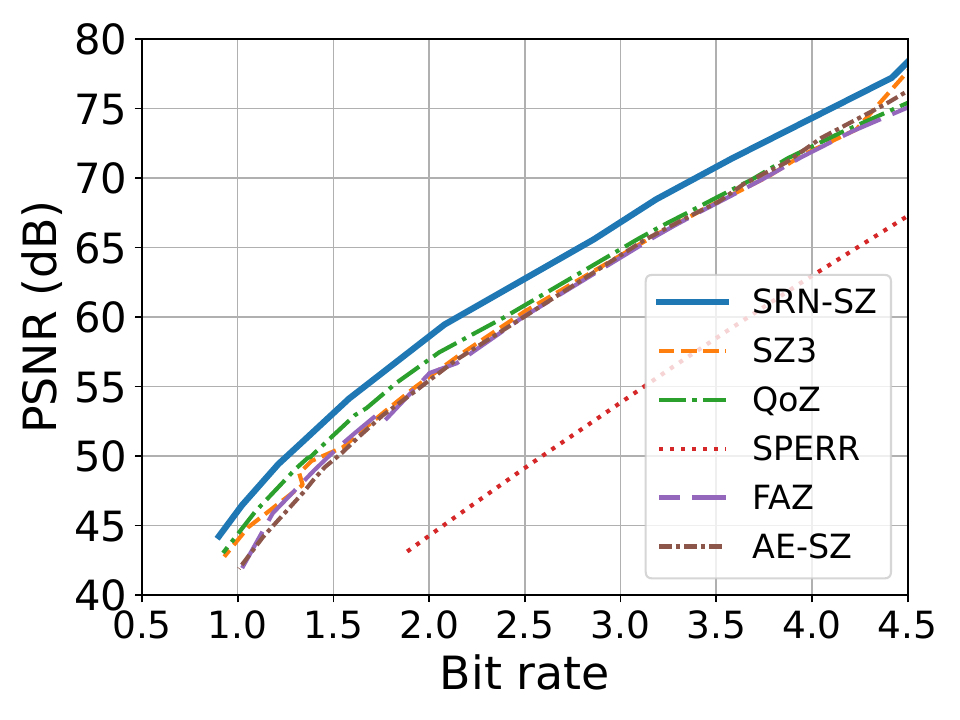}}%
}
\hspace{-6mm}
\subfigure[{Hurricane-QGRAUP}]
{
\raisebox{-1cm}{\includegraphics[scale=0.26]{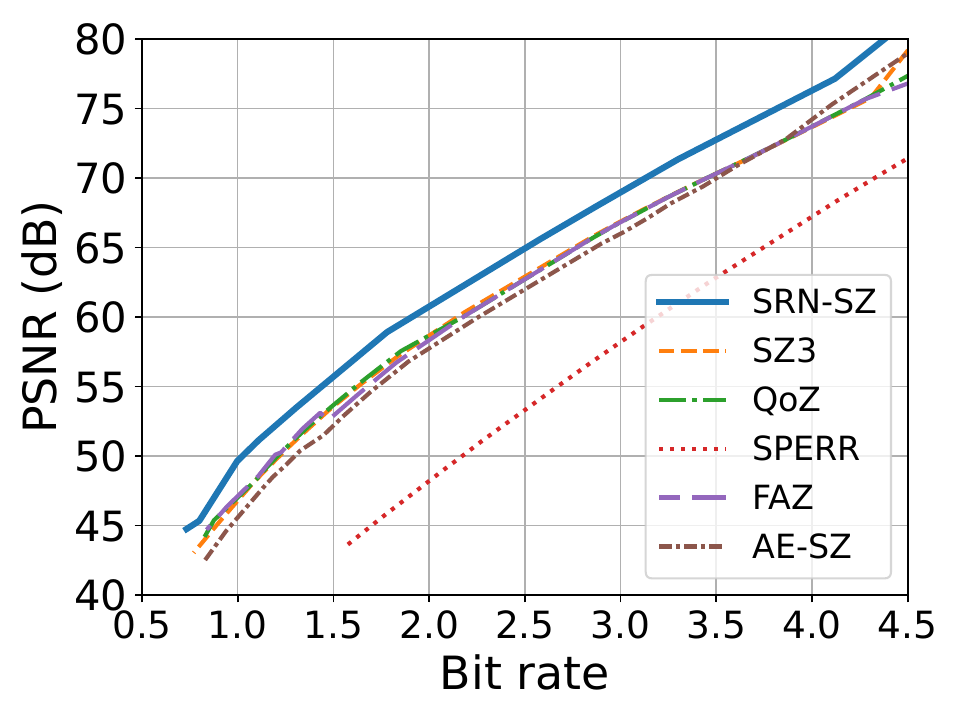}}%
}
\hspace{-10mm}
\vspace{-1mm}

\caption{Rate Distortion Evaluation (PSNR)}
\label{fig:rate-psnr}
\end{figure}

Figure \ref{fig:rate-psnr} displays the rate-distortion evaluation results of each lossy compressor on all datasets. In the plots, the x-axis is bit rate and the y-axis is PSNR. Like the cases of same-error-bound compression ratios, SRN-SZ has the best rate-distortion curves on all the datasets. On the CESM-CLDHGH dataset, SRN-SZ achieves $60\%$ to $80\%$ compression ratio improvement than the second-best SPERR in the PSNR range of 70 $\sim$ 80. On the Ocean-TMXL dataset, SRN-SZ achieves $\sim$20\% compression ratio improvement than the second-best QoZ in the PSNR range of 60 $\sim$ 70. Additionally, SRN-SZ overperforms all other compressors by about $5\%$ to $15\%$  compression ratio improvements on the rest of the datasets. 

Those results show that, for certain datasets on which the traditional or autoencoder-based lossy compressors can only present limited compression ratios, SRN-SZ has the potential to optimize the compression of those datasets to a further extent, and the reasons can be attributed to 3 terms. First, those datasets have complex data characteristics and patterns for which traditional data modeling techniques cannot fit well. Second, the newly proposed compression framework of SRN-SZ enables the compressor to directly leverage a super-resolution network for the data prediction via data grid expansion (super-resolution) instead of applying a redundant autoencoder model for which the latent vectors must be stored (such as AE-SZ does). Third, the hybrid usage of interpolations and super-resolution networks makes the interpolation compensate for the limitation of neural networks when dealing with small data grids.

\subsubsection{Visualization of decompressed data}

\begin{figure}[ht] \centering

\hspace{-8mm}
\subfigure[{CESM-CLDHGH (Original)}]
{
\raisebox{-1cm}{\includegraphics[scale=0.31]{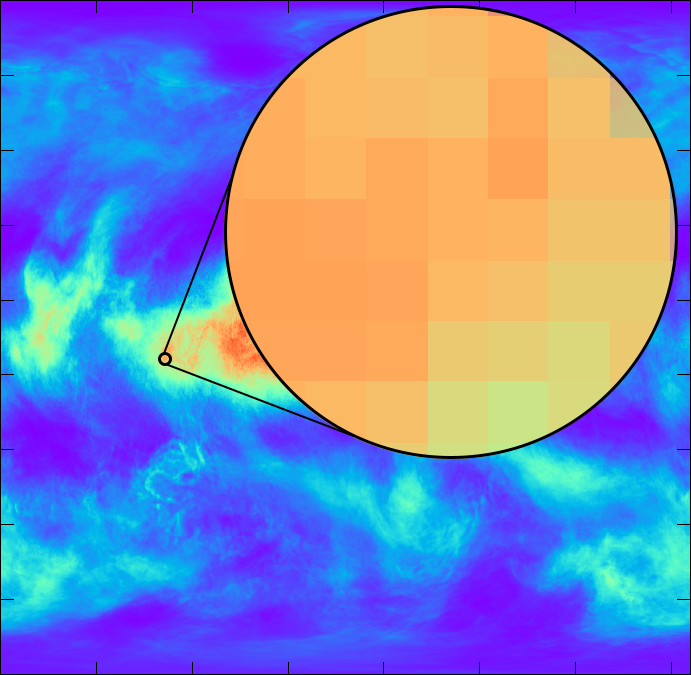}}
}
\hspace{-3mm}
\subfigure[{SRN-SZ (PSNR:68.5,CR:31.6)}]
{
\raisebox{-1cm}{\includegraphics[scale=0.31]{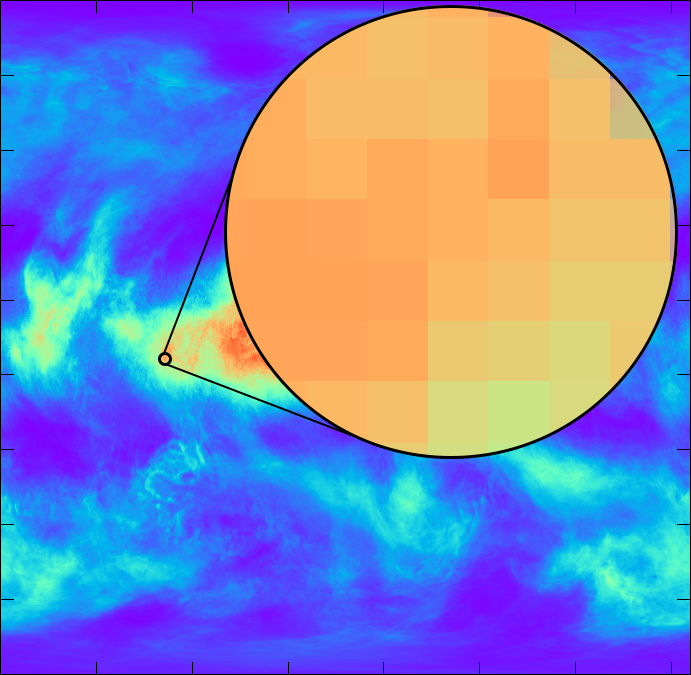}}
}
\hspace{-10mm}

\hspace{-7mm}
\subfigure[{SZ3 (PSNR:59.7,CR:31.9)}]
{
\raisebox{-1cm}{\includegraphics[scale=0.31]{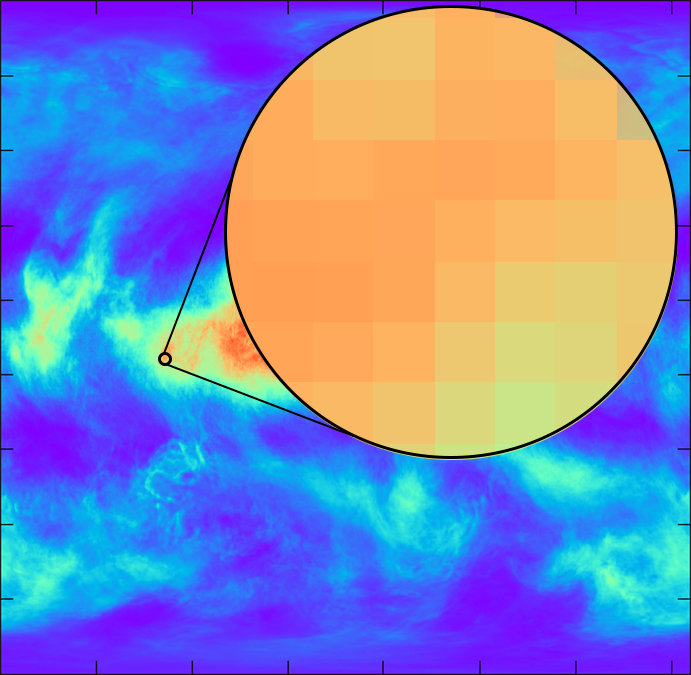}}
}
\hspace{-3mm}
\subfigure[{QoZ (PSNR:60.9,CR:31.6)}]
{
\raisebox{-1cm}{\includegraphics[scale=0.31]{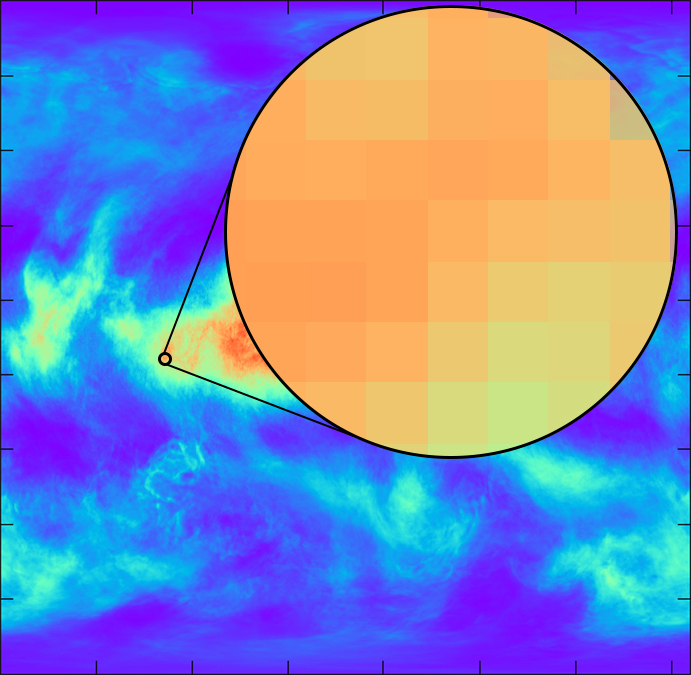}}
}
\hspace{-10mm}

\hspace{-7mm}
\subfigure[{SPERR (PSNR:61.8,CR:31.9)}]
{
\raisebox{-1cm}{\includegraphics[scale=0.31]{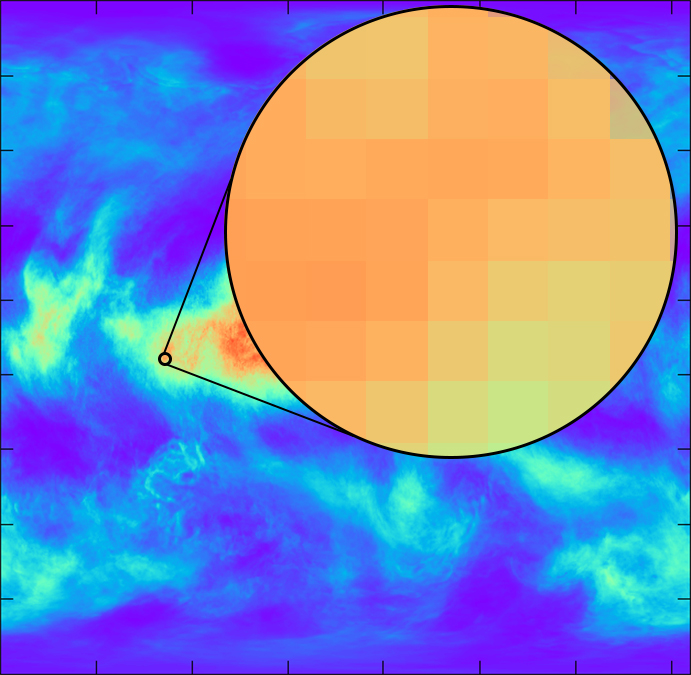}}
}
\hspace{-3mm}
\subfigure[{FAZ (PSNR:63.4,CR:31.4)}]
{
\raisebox{-1cm}{\includegraphics[scale=0.31]{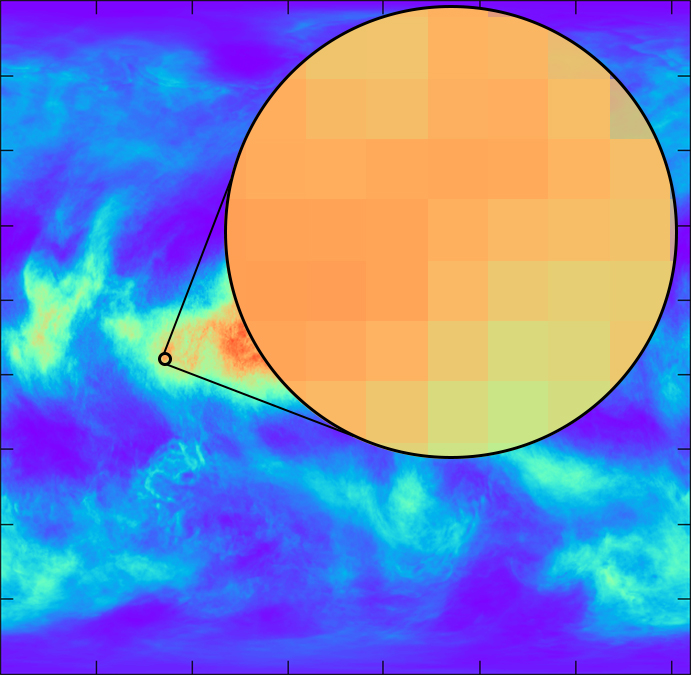}}
}
\hspace{-10mm}
\vspace{-3mm}

\caption{Visualization of reconstructed data (CESM-CLDHGH)}
\label{fig:vis}
\end{figure}

As an example of the high compression quality of SRN-SZ, In Figure \ref{fig:vis}, we present several visualizations of the decompression results of CESM-CLDHGH data field from multiple compressors, together with the original data as the reference. For a fair comparison, for each compressor, the data are compressed into a fixed compression ratio (around 32) and then get decompressed. According to Figure \ref{fig:vis} (we omit the visualization results of AE-SZ because it has poor visualization quality with PSNR $\approx$ 53 under the specified compression ratio), in this case, the decompression data of SRN-SZ has the lowest distortion from the original input, with a PSNR of 68.5 which is 5 higher than the second-best FAZ. The zoomed regions also show that SRN-SZ has best preserved the local data patterns as well. The local visualization of SRN-SZ decompressed data is nearly identical to the original data, meanwhile, the ones of other compressors suffer from some quality degradation.

\subsubsection{Ablation Study}
\label{sec:abla}

For verifying and understanding how the design details of SRN-SZ contribute to the overall compression quality, especially for the design components in the network pre-training pipelines, we conduct several ablation studies for the network pre-training, identifying and quantifying the contributions of the corresponding design components.

\begin{figure}[ht] 
\centering
\hspace{-10mm}
\subfigure[{NYX-Dark Matter Density}]
{
\raisebox{-1cm}{\includegraphics[scale=0.25]{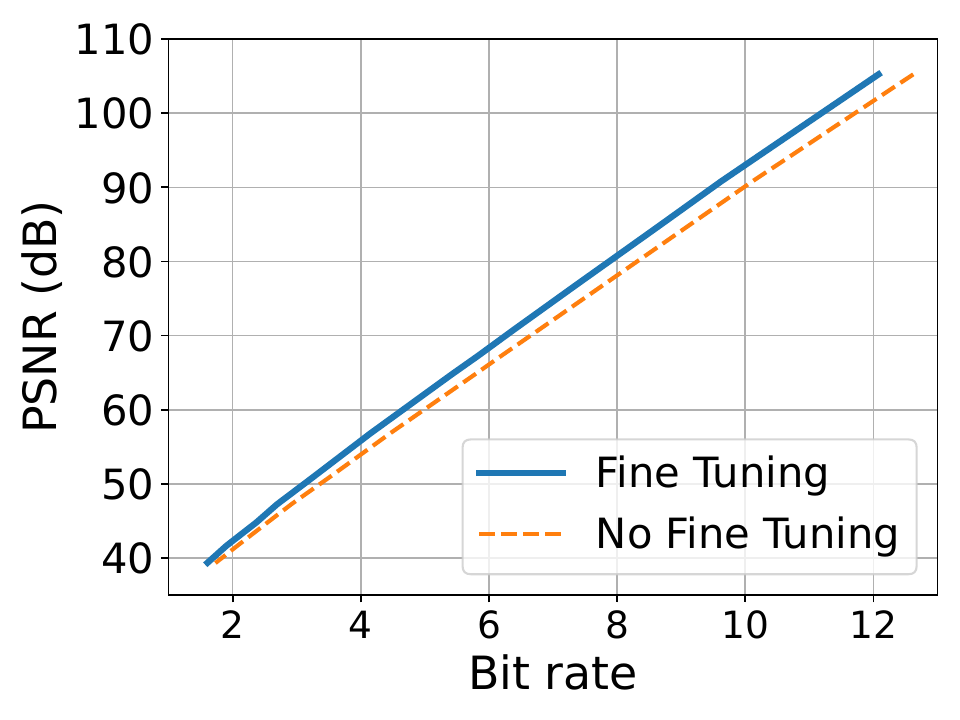}}%
}
\hspace{-4mm}
\subfigure[{Ocean-TMXL}]
{
\raisebox{-1cm}{\includegraphics[scale=0.25]{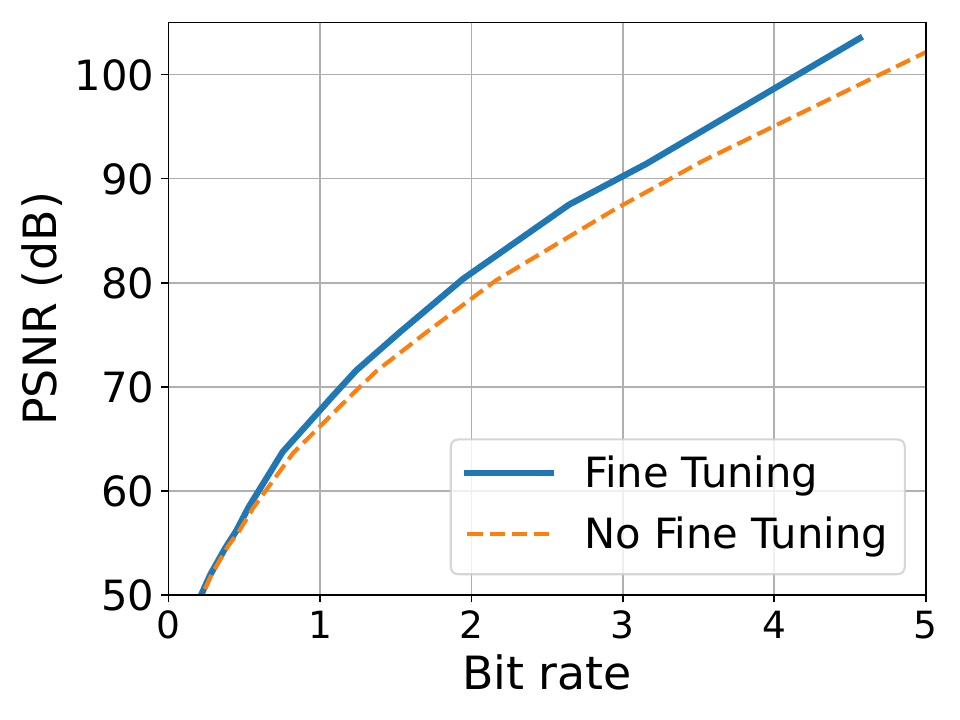}}%
}
\hspace{-10mm}
\vspace{-1mm}
\caption{Ablation study for the Domain-specific fine-tuning}
\label{fig:ab-ft}
\end{figure}

First, to examine the impact of domain-specific fine-tuning (described in Section \ref{sec:ft}) on the training of HAT networks in SRN-SZ. We tested the compression of SRN-SZ with networks free of domain-specific fine-tuning and then compared the rate-distortion of it to the one from ordinary SRN-SZ. This comparison is detailed in Figure \ref{fig:ab-ft} with 2 examples presented (on Ocean-TMXL and NYX-Dark Matter Density). It is shown that the domain-specific fine-tuning process (the blue curves in Figure \ref{fig:ab-ft}) can consistently improve the compression rate distortion over the SRN-SZ without a network fine-tuning process (the orange curves in Figure \ref{fig:ab-ft}).

\begin{figure}[ht] 
\centering
\hspace{-10mm}
\subfigure[{CESM-CLDHGH}]
{
\raisebox{-1cm}{\includegraphics[scale=0.25]{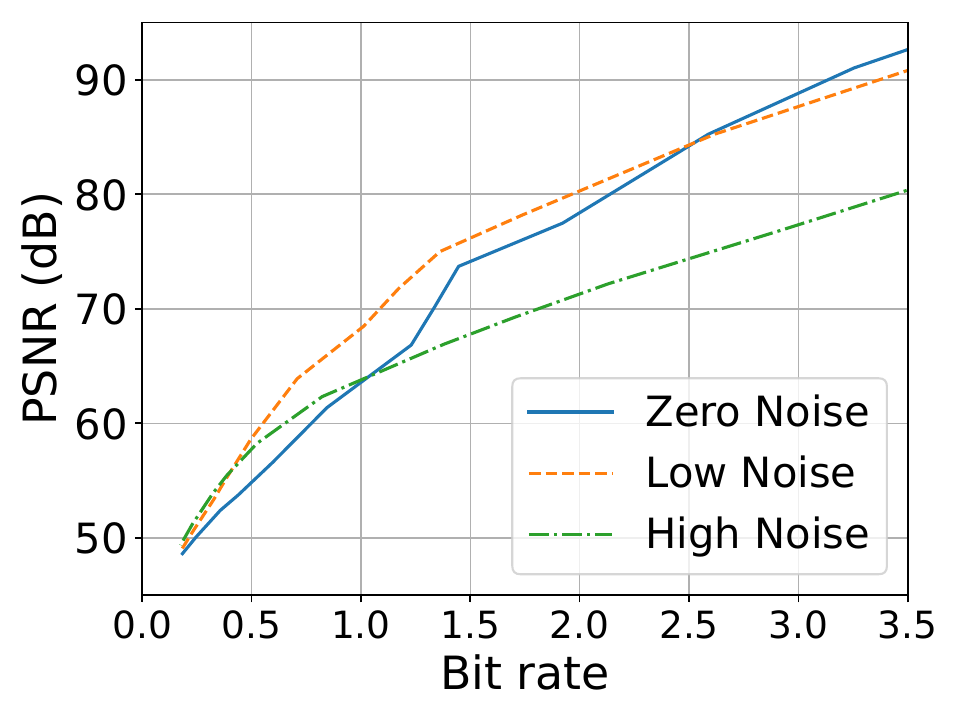}}%
}
\hspace{-4mm}
\subfigure[{CESM-FREQSH}]
{
\raisebox{-1cm}{\includegraphics[scale=0.25]{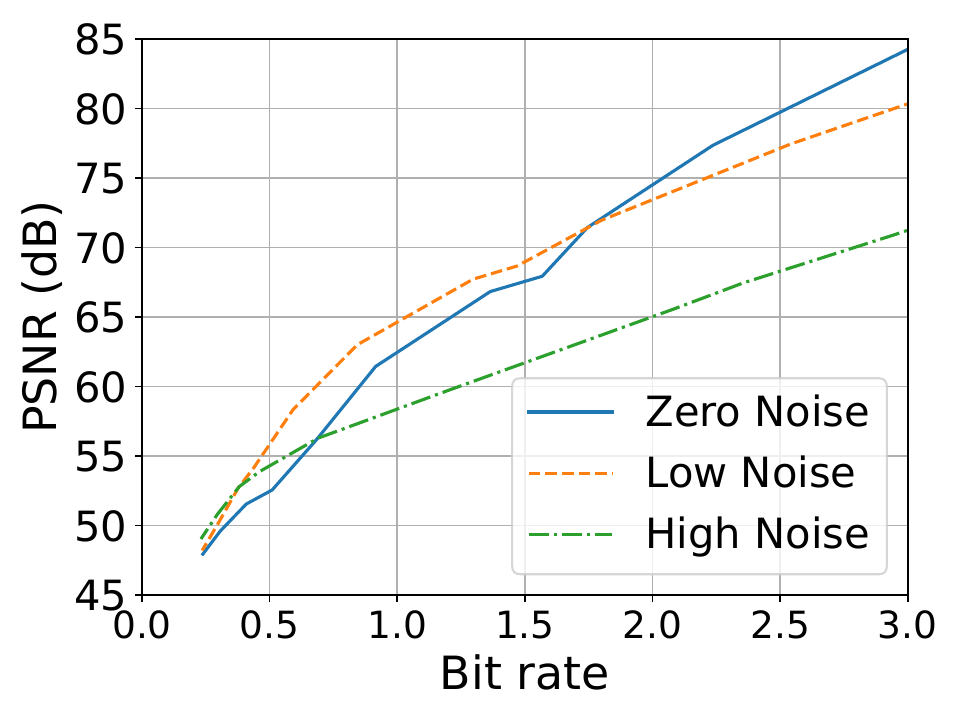}}%
}
\hspace{-10mm}
\vspace{-1mm}

\caption{Ablation study for the Denoise Training}
\label{fig:ab-noise}
\end{figure}

Next, we address the importance of SRN-SZ denoise training via analyzing and comparing the compression rate-distortion of SRN-SZ integrating fixed HAT networks each trained with a certain intensity of noise added to the training data. In Figure \ref{fig:ab-noise}, the rate-PSNR curves of SRN-SZ with HAT networks trained by 3 different levels of noise intensity (zero noise, low noise of $\sigma$=1e-3, and high noise of $\sigma$=1e-2) are illustrated. Those compressors exhibit advantages over the others on different bit rate ranges. SRN-SZ with high-noise-trained overperforms the other configurations when the bit rate is smaller than 0.4 (corresponding to error bound $>$ 1e-2). The low-noise-trained HAT network optimizes the SRN-SZ compression under medium bit rates, and when the bit rate is large (error bound $<$ 1e-4), Leveraging networks trained with no noise achieves the best rate-distortion. From those results, we prove that the error-bound-adaptive dynamic usage of differently-trained HAT networks (with diverse noise intensities) essentially optimizes the compression of SRN-SZ.

\section{Conclusion and Future Work}
\label{sec:conclusion}

In this paper, We propose SRN-SZ, a deep learning-based error-bounded compressor that leverages one of the most advanced super-resolution neural network archetypes, namely HAT. SRN-SZ abstracts the data prediction process in compression into a hierarchical data grid expansion paradigm, enabling the utility of super-resolution neural networks for lossy compression. To exploit the advantages of different data reconstruction techniques, the data grid expansion in SRN-SZ is performed by a self-adaptive hybrid method of super-resolution HAT networks and interpolations. For the better adaptation of super-resolution networks to scientific data, SRN-SZ integrates a carefully designed network training pipeline for optimizing the network performance. In the evaluations, SRN-SZ outperforms all other state-of-the-art error-bounded lossy compressors in terms of compression ratio and rate-distortion, achieving up to 75\% compression ratio improvements under the same error bound and up to 80\% compression ratio improvements under the same PSNR.

SRN-SZ still has a few limitations. First, since it is based on neural networks, its running speed is inevitably quite lower than traditional lossy compressors, and the complexity of its integrated network makes it slower than some autoencoder-based compressors such as AE-SZ. Second, the compression ratios of SRN-SZ may not outperform the existing state-of-the-art compressors on datasets with high compressibility. Third, the training of HAT networks in SRN-SZ is not fully optimized. In future work, we will revise SRN-SZ in several aspects such as accelerating and fine-tuning the training and inference of its integrated neural networks, improving its compression ratio on easy-to-compress datasets, and so on.

\section*{Acknowledgments}
This research was supported by the Exascale Computing Project (ECP), Project Number: 17-SC-20-SC, a collaborative effort of two DOE organizations – the Office of Science and the National Nuclear Security Administration, responsible for the planning and preparation of a capable exascale ecosystem, including software, applications, hardware, advanced system engineering and early testbed platforms, to support the nation’s exascale computing imperative. The material was supported by the U.S. Department of Energy, Office of Science, Advanced Scientific Computing Research (ASCR), under contract DE-AC02-06CH11357, and supported by the National Science Foundation under Grant OAC-2003709, OAC-2104023, OAC-2311875,  OAC-2311877, and OAC-2153451. We acknowledge the computing resources provided on Bebop (operated by Laboratory Computing Resource Center at Argonne) and on Theta and JLSE (operated by Argonne Leadership Computing Facility).

\bibliographystyle{IEEEtran}
\bibliography{references}

\begin{thebibliography}{10}
\providecommand{\url}[1]{#1}
\csname url@samestyle\endcsname
\providecommand{\newblock}{\relax}
\providecommand{\bibinfo}[2]{#2}
\providecommand{\BIBentrySTDinterwordspacing}{\spaceskip=0pt\relax}
\providecommand{\BIBentryALTinterwordstretchfactor}{4}
\providecommand{\BIBentryALTinterwordspacing}{\spaceskip=\fontdimen2\font plus
\BIBentryALTinterwordstretchfactor\fontdimen3\font minus \fontdimen4\font\relax}
\providecommand{\BIBforeignlanguage}[2]{{%
\expandafter\ifx\csname l@#1\endcsname\relax
\typeout{** WARNING: IEEEtran.bst: No hyphenation pattern has been}%
\typeout{** loaded for the language `#1'. Using the pattern for}%
\typeout{** the default language instead.}%
\else
\language=\csname l@#1\endcsname
\fi
#2}}
\providecommand{\BIBdecl}{\relax}
\BIBdecl

\bibitem{szinterp}
K.~Zhao, S.~Di, M.~Dmitriev, T.-L.~D. Tonellot, Z.~Chen, and F.~Cappello, ``Optimizing error-bounded lossy compression for scientific data by dynamic spline interpolation,'' in \emph{2021 IEEE 37th International Conference on Data Engineering (ICDE)}, 2021, pp. 1643--1654.

\bibitem{sz3}
X.~Liang, K.~Zhao, S.~Di, S.~Li, R.~Underwood, A.~M. Gok, J.~Tian, J.~Deng, J.~C. Calhoun, D.~Tao \emph{et~al.}, ``{SZ3}: A modular framework for composing prediction-based error-bounded lossy compressors,'' \emph{IEEE Transactions on Big Data}, 2022.

\bibitem{zfp}
P.~Lindstrom, ``Fixed-rate compressed floating-point arrays,'' \emph{IEEE transactions on visualization and computer graphics}, vol.~20, no.~12, pp. 2674--2683, 2014.

\bibitem{SPERR}
S.~Li, P.~Lindstrom, and J.~Clyne, ``Lossy scientific data compression with sperr,'' in \emph{2023 IEEE International Parallel and Distributed Processing Symposium (IPDPS)}.\hskip 1em plus 0.5em minus 0.4em\relax IEEE, 2023, pp. 1007--1017.

\bibitem{ae-sz}
J.~Liu, S.~Di, K.~Zhao, S.~Jin, D.~Tao, X.~Liang, Z.~Chen, and F.~Cappello, ``Exploring autoencoder-based error-bounded compression for scientific data,'' in \emph{2021 IEEE International Conference on Cluster Computing (CLUSTER)}.\hskip 1em plus 0.5em minus 0.4em\relax IEEE, 2021, pp. 294--306.

\bibitem{han2022coordnet}
J.~Han and C.~Wang, ``Coordnet: Data generation and visualization generation for time-varying volumes via a coordinate-based neural network,'' \emph{IEEE Transactions on Visualization and Computer Graphics}, 2022.

\bibitem{huang2022compressing}
L.~Huang and T.~Hoefler, ``Compressing multidimensional weather and climate data into neural networks,'' \emph{arXiv preprint arXiv:2210.12538}, 2022.

\bibitem{lu2021compressive}
Y.~Lu, K.~Jiang, J.~A. Levine, and M.~Berger, ``Compressive neural representations of volumetric scalar fields,'' in \emph{Computer Graphics Forum}, vol.~40, no.~3.\hskip 1em plus 0.5em minus 0.4em\relax Wiley Online Library, 2021, pp. 135--146.

\bibitem{hayne2021using}
L.~Hayne, J.~Clyne, and S.~Li, ``Using neural networks for two dimensional scientific data compression,'' in \emph{2021 IEEE International Conference on Big Data (Big Data)}.\hskip 1em plus 0.5em minus 0.4em\relax IEEE, 2021, pp. 2956--2965.

\bibitem{Xin-bigdata18}
X.~Liang, S.~Di, D.~Tao, S.~Li, S.~Li, H.~Guo, Z.~Chen, and F.~Cappello, ``Error-controlled lossy compression optimized for high compression ratios of scientific datasets,'' in \emph{2018 {IEEE} International Conference on Big Data}.\hskip 1em plus 0.5em minus 0.4em\relax IEEE, 2018.

\bibitem{qoz}
J.~Liu, S.~Di, K.~Zhao, X.~Liang, Z.~Chen, and F.~Cappello, ``Dynamic quality metric oriented error bounded lossy compression for scientific datasets,'' in \emph{2022 SC22: International Conference for High Performance Computing, Networking, Storage and Analysis (SC)}.\hskip 1em plus 0.5em minus 0.4em\relax IEEE Computer Society, 2022, pp. 892--906.

\bibitem{ballester2019tthresh}
R.~Ballester-Ripoll, P.~Lindstrom, and R.~Pajarola, ``{TTHRESH}: Tensor compression for multidimensional visual data,'' \emph{IEEE transactions on visualization and computer graphics}, vol.~26, no.~9, pp. 2891--2903, 2019.

\bibitem{cusz}
J.~Tian \emph{et~al.}, ``{CuSZ}: An efficient gpu-based error-bounded lossy compression framework for scientific data,'' in \emph{Proceedings of the ACM International Conference on Parallel Architectures and Compilation Techniques}, ser. PACT '20, 2020, p. 3–15.

\bibitem{cusz+}
J.~Tian, S.~Di, X.~Yu, C.~Rivera, K.~Zhao, S.~Jin, Y.~Feng, X.~Liang, D.~Tao, and F.~Cappello, ``cusz (x): Optimizing error-bounded lossy compression for scientific data on gpus.'' \emph{CoRR}, 2021.

\bibitem{FZGPU}
B.~Zhang, J.~Tian, S.~Di, X.~Yu, Y.~Feng, X.~Liang, D.~Tao, and F.~Cappello, ``Fz-gpu: A fast and high-ratio lossy compressor for scientific computing applications on gpus,'' \emph{arXiv preprint arXiv:2304.12557}, 2023.

\bibitem{dong2014learning}
C.~Dong, C.~C. Loy, K.~He, and X.~Tang, ``Learning a deep convolutional network for image super-resolution,'' in \emph{Computer Vision--ECCV 2014: 13th European Conference, Zurich, Switzerland, September 6-12, 2014, Proceedings, Part IV 13}.\hskip 1em plus 0.5em minus 0.4em\relax Springer, 2014, pp. 184--199.

\bibitem{ahn2018fast}
N.~Ahn, B.~Kang, and K.-A. Sohn, ``Fast, accurate, and lightweight super-resolution with cascading residual network,'' in \emph{Proceedings of the European conference on computer vision (ECCV)}, 2018, pp. 252--268.

\bibitem{lim2017enhanced}
B.~Lim, S.~Son, H.~Kim, S.~Nah, and K.~Mu~Lee, ``Enhanced deep residual networks for single image super-resolution,'' in \emph{Proceedings of the IEEE conference on computer vision and pattern recognition workshops}, 2017, pp. 136--144.

\bibitem{zhang2018image}
Y.~Zhang, K.~Li, K.~Li, L.~Wang, B.~Zhong, and Y.~Fu, ``Image super-resolution using very deep residual channel attention networks,'' in \emph{Proceedings of the European conference on computer vision (ECCV)}, 2018, pp. 286--301.

\bibitem{transformer}
A.~Vaswani, N.~Shazeer, N.~Parmar, J.~Uszkoreit, L.~Jones, A.~N. Gomez, {\L}.~Kaiser, and I.~Polosukhin, ``Attention is all you need,'' \emph{Advances in neural information processing systems}, vol.~30, 2017.

\bibitem{vit}
A.~Dosovitskiy, L.~Beyer, A.~Kolesnikov, D.~Weissenborn, X.~Zhai, T.~Unterthiner, M.~Dehghani, M.~Minderer, G.~Heigold, S.~Gelly \emph{et~al.}, ``An image is worth 16x16 words: Transformers for image recognition at scale,'' \emph{arXiv preprint arXiv:2010.11929}, 2020.

\bibitem{liu2021swin}
Z.~Liu, Y.~Lin, Y.~Cao, H.~Hu, Y.~Wei, Z.~Zhang, S.~Lin, and B.~Guo, ``Swin transformer: Hierarchical vision transformer using shifted windows,'' in \emph{Proceedings of the IEEE/CVF international conference on computer vision}, 2021, pp. 10\,012--10\,022.

\bibitem{wu2021cvt}
H.~Wu, B.~Xiao, N.~Codella, M.~Liu, X.~Dai, L.~Yuan, and L.~Zhang, ``Cvt: Introducing convolutions to vision transformers,'' in \emph{Proceedings of the IEEE/CVF international conference on computer vision}, 2021, pp. 22--31.

\bibitem{chen2021pre}
H.~Chen, Y.~Wang, T.~Guo, C.~Xu, Y.~Deng, Z.~Liu, S.~Ma, C.~Xu, C.~Xu, and W.~Gao, ``Pre-trained image processing transformer,'' in \emph{Proceedings of the IEEE/CVF conference on computer vision and pattern recognition}, 2021, pp. 12\,299--12\,310.

\bibitem{liang2021swinir}
J.~Liang, J.~Cao, G.~Sun, K.~Zhang, L.~Van~Gool, and R.~Timofte, ``Swinir: Image restoration using swin transformer,'' in \emph{Proceedings of the IEEE/CVF international conference on computer vision}, 2021, pp. 1833--1844.

\bibitem{hat}
X.~Chen, X.~Wang, J.~Zhou, Y.~Qiao, and C.~Dong, ``Activating more pixels in image super-resolution transformer,'' in \emph{Proceedings of the IEEE/CVF Conference on Computer Vision and Pattern Recognition}, 2023, pp. 22\,367--22\,377.

\bibitem{miranda}
\BIBentryALTinterwordspacing
Miranda application. [Online]. Available: \url{https://wci.llnl.gov/simulation/computer-codes/miranda}
\BIBentrySTDinterwordspacing

\bibitem{liu2021high}
T.~Liu, J.~Wang, Q.~Liu, S.~Alibhai, T.~Lu, and X.~He, ``High-ratio lossy compression: Exploring the autoencoder to compress scientific data,'' \emph{IEEE Transactions on Big Data}, 2021.

\bibitem{MGARD}
M.~Ainsworth, O.~Tugluk, B.~Whitney, and S.~Klasky, ``Multilevel techniques for compression and reduction of scientific data—the univariate case,'' \emph{Computing and Visualization in Science}, vol.~19, no.~5, pp. 65--76, 2018.

\bibitem{liang2021mgard+}
X.~Liang, B.~Whitney, J.~Chen, L.~Wan, Q.~Liu, D.~Tao, J.~Kress, D.~R. Pugmire, M.~Wolf, N.~Podhorszki \emph{et~al.}, ``Mgard+: Optimizing multilevel methods for error-bounded scientific data reduction,'' \emph{IEEE Transactions on Computers}, 2021.

\bibitem{he2016deep}
K.~He, X.~Zhang, S.~Ren, and J.~Sun, ``Deep residual learning for image recognition,'' in \emph{Proceedings of the IEEE conference on computer vision and pattern recognition}, 2016, pp. 770--778.

\bibitem{shi2016real}
W.~Shi, J.~Caballero, F.~Husz{\'a}r, J.~Totz, A.~P. Aitken, R.~Bishop, D.~Rueckert, and Z.~Wang, ``Real-time single image and video super-resolution using an efficient sub-pixel convolutional neural network,'' in \emph{Proceedings of the IEEE conference on computer vision and pattern recognition}, 2016, pp. 1874--1883.

\bibitem{HPEZ}
``{HPEZ},'' \url{https://github.com/Meso272/HPEZ.git}.

\bibitem{cesm}
J.~E. Kay and et\ al., ``The {Community Earth System Model (CESM)} large ensemble project: A community resource for studying climate change in the presence of internal climate variability,'' \emph{Bulletin of the American Meteorological Society}, vol.~96, no.~8, pp. 1333--1349, 2015.

\bibitem{geodriveFirstBreak2020}
S.~Kayum \emph{et~al.}, ``{GeoDRIVE} -- a high performance computing flexible platform for seismic applications,'' \emph{First Break}, vol.~38, no.~2, pp. 97--100, 2020.

\bibitem{jhtdb}
Y.~Li, E.~Perlman, M.~Wan, Y.~Yang, C.~Meneveau, R.~Burns, S.~Chen, A.~Szalay, and G.~Eyink, ``A public turbulence database cluster and applications to study lagrangian evolution of velocity increments in turbulence,'' \emph{Journal of Turbulence}, no.~9, p. N31, 2008.

\bibitem{hurricane}
{Hurricane ISABEL simulation data}, \url{http://vis.computer.org/vis2004contest/data.html}, 2004, online.

\bibitem{scale-letkf}
``Scalable computing for advanced library and environment (scale) -- letkf,'' \url{https://github.com/gylien/scale-letkf}.

\bibitem{nyx}
{NYX simulation}, \url{https://amrex-astro.github.io/Nyx}, 2019, online.

\bibitem{liu2023faz}
J.~Liu, S.~Di, K.~Zhao, X.~Liang, Z.~Chen, and F.~Cappello, ``Faz: A flexible auto-tuned modular error-bounded compression framework for scientific data,'' in \emph{Proceedings of the 37th International Conference on Supercomputing}, 2023, pp. 1--13.

\bibitem{sdrb}
K.~Zhao, S.~Di, X.~Lian, S.~Li, D.~Tao, J.~Bessac, Z.~Chen, and F.~Cappello, ``{SDRBench}: Scientific data reduction benchmark for lossy compressors,'' in \emph{2020 IEEE International Conference on Big Data (Big Data)}, 2020, pp. 2716--2724.

\end{thebibliography}

\end{document}